# Extended Triangular Method: A Generalized Algorithm for Contradiction Separation Based Automated Deduction

Yang Xu[a, b*], Shuwei Chen[a, b*], Jun Liu[b, c], Feng Cao[c, d], and Xingxing He[a, b]

[a] *School of Mathematics, Southwest Jiaotong University, Chengdu 611756, China*
[b] *National-Local Joint Engineering Laboratory of System Credibility Automatic Verification, Southwest Jiaotong University, Chengdu 611756, China*
[c] *School of Computing, Ulster University, Belfast BT15 1ED, Northern Ireland, UK*
[d] *School of Information Engineering, Jiangxi University of Science and Technology, Ganzhou 341000, China*

**Abstract:** Automated deduction lies at the core of Artificial Intelligence (AI), underpinning theorem proving, formal verification, and logical reasoning. Despite decades of progress, reconciling deductive completeness with computational efficiency remains an enduring challenge. Traditional reasoning calculi, grounded in binary resolution, restrict inference to pairwise clause interactions and thereby limit deductive synergy among multiple clauses. The Contradiction Separation Extension (CSE) framework, introduced in 2018, proposed a dynamic multi-clause reasoning theory that redefined logical inference as a process of *contradiction separation* rather than sequential resolution. While that work established the theoretical foundation, its algorithmic realization remained unformalized and unpublished. This work presents the Extended Triangular Method (ETM), a generalized contradiction-construction algorithm that formalizes and extends the internal mechanisms of contradiction separation. The ETM unifies multiple contradiction-building strategies, including the earlier *Standard Extension* method, within a triangular geometric framework that supports flexible clause interaction and dynamic synergy. ETM serves as the algorithmic core of several high-performance theorem provers, CSE, CSE_E, CSI_E, and CSI_Enig, whose competitive results in standard first-order benchmarks (TPTP problem sets and CASC 2018–2025; Sutcliffe, 2025) empirically validate the effectiveness and generality of the proposed approach. By bridging theoretical abstraction and operational implementation, ETM advances the contradiction separation paradigm into a generalized, scalable, and practically competitive model for automated reasoning, offering new directions for future research in logical inference and theorem proving.

## 1. Introduction

Automated deduction, as a core branch of artificial intelligence and formal logic, has long aimed to enable

[*] Corresponding authors.
All the authors are co-first authors.
*E-mail addresses*: xuyang@swjtu.edu.cn (Y. Xu), swchen@swjtu.edu.cn (S. Chen), j.liu@ulster.ac.uk (J. Liu), caofeng19840301@163.com (F. Cao), x.he@swjtu.edu.cn (X. He)

machines to perform logical reasoning autonomously and reliably (Russell & Norvig, 2021). Since the introduction of the *resolution principle* by Robinson (1965), resolution-based theorem proving has played a dominant role in automated reasoning for over five decades. Numerous refinements, restrictions, and strategies have been developed to improve the efficiency of resolution-based automated deduction systems, including set-of-support, semantic, linear, hyper-, and unit-resulting resolutions, among others. However, despite their success, these methods fundamentally rely on *binary resolution*, wherein contradictions are derived through pairwise clause interactions and each inference step involves only two clauses. This syntactic reliance on binary operations inherently limits deductive flexibility and restricts the exploitation of synergistic interactions among multiple clauses, often leading to redundant derivations and exponential search growth in large-scale problems, such as those found in the Thousands of Problems for Theorem Provers (TPTP) benchmark library (Sutcliffe, 2017).

To address the inherent restrictiveness of binary resolution, Xu *et al*. (2018) proposed the *Contradiction Separation (CS) based Dynamic Multi-Clause Synergized Automated Deduction* framework, denoted as **Contradiction Separation Extension (CSE)**, the E was added to emphasize that the framework extends the basic CS idea into a dynamic, multi-clause, synergized deduction theory. In CSE, contradictions are no longer viewed as complementary pairs between two clauses, but as general unsatisfiable clause sets consisting of multiple clauses working cooperatively in the deduction process. The resulting *CS rule* enables deduction steps involving multiple clauses dynamically and synergistically, thereby generalizing binary resolution as a special case. The 2018 paper proved soundness and completeness, defining a broad logical framework that has since become the theoretical basis for multiple theorem-proving systems.

However, the 2018 CSE paper was deliberately conceptual, it focused on the logical theory of contradiction separation, without disclosing the specific algorithmic process for constructing contradictions and managing multi-clause interactions during proof search. The concrete algorithmic realization of contradiction separation, particularly how contradictions are incrementally constructed, maintained, and extended, remained unpublished. This algorithmic layer became the algorithmic kernel of subsequent implementations (CASC 2018-2015; Sutcliffe, 2025) such as CSE, CSE_E, CSI_E, and CSI_Enig, which all used the internal contradiction-construction procedures derived from this method but did not document them in formal literature.

This work closes that gap by introducing the Extended Triangular Contradiction Separation Based Dynamic Automated Deduction Method (hereafter *Extended Triangular Method*, ETM). The method is motivated by the observation that efficient deduction depends not only on the underlying inference rule but also on how contradictions are constructed and exploited during proof search. The ETM explicitly defines the complete algorithmic realization of contradiction construction within the CSE framework. ETM formalizes and extends the internal mechanisms of contradiction separation through a triangular geometric inference model that supports flexible, multi-directional clause interaction. In contrast to binary resolution, where inference is restricted to two clauses, ETM constructs contradictions among three or more interacting clauses in a dynamically expanding structure. While the Standard Extension (SE) method (Xu *et al*., 2025b), as an independent contradiction-construction strategy, used fixed complementary-literal expansion, i.e., complementary-literal-based clause selection, the ETM generalizes and unifies this to dynamic, goal-directed selection of clauses and literals, thereby operationalizing the theoretical CSE framework.

Beyond theoretical generalization, the ETM thus serves as the algorithmic bridge between theory and implementation: it translates the abstract inference principles of CSE into an executable procedure that forms algorithmic kernel for a family of high-performance theorem provers (CASC 2018-2025; Sutcliffe, 2025), including CSE, CSE_E, CSI_E, and CSI_Enig, which have achieved top-tier results in international automated reasoning competitions such as CASC (2018–2025). Its publication will complete the conceptual–practical cycle of the CSE research line. Moreover, as an inference mechanism, it can be integrated with other automated reasoning systems, including those built on the CSE_E, CSI_E, and CSI_Enig architectures (see Section 7 for details), enabling hybrid or learning-guided versions of the proposed method.

The main contributions of the present work are summarized as follows:

1) **Algorithmic Disclosure and Formalization:**

    The paper presents the explicit **construction algorithm** of contradiction separation reasoning, the algorithmic basis that has powered subsequent systems derived from the 2018 CSE theory.

2) **Extension of the Standard Triangular Method:**

    The proposed ETM generalizes the fixed-literal triangular construction into a dynamic, goal-oriented mechanism, improving flexibility, adaptability, and deduction efficiency.

3) **Unification of Theory and Implementation:**

    The paper links the 2018 theoretical framework with practical inference realization, providing the missing algorithmic layer that connects CSE theory to the CSE, CSE_E, CSI_E, and CSI_Enig implementations.

4) **Soundness, Completeness, and Generality:**

    Rigorous proofs establish that the new algorithm preserves the soundness and completeness of CSE while supporting more efficient multi-clause synergy.

5) **Integration with Hybrid and Learning-Enhanced Provers:**

    The method can be integrated with advanced reasoning frameworks such as CSE_E, CSI_E, and CSI_Enig, enabling hybrid and learning-guided automated reasoning.

6) **Empirical Validation through Proven Implementations:**

    Instead of performing redundant experiments, the paper analyzes the competition-proven performance of CSE, CSE_E, CSI_E, and CSI_Enig, systems that implicitly employ the proposed algorithm, providing indirect yet strong empirical validation.

Through these contributions, ETM establishes a new foundation for dynamic, structure-aware automated deduction that is both theoretically principled and practically effective.

The remainder of this paper is structured as follows. Section 2 reviews related works. Section 3 outlines preliminaries and basic concepts of contradiction separation based automated deduction. Section 4 and Section 5 present the Extended Triangular method in propositional and first-order logic, respectively, including definitions, algorithms and formal proofs of soundness and completeness.. Section 6 discusses the relationship between the ETM and resolution based inference method, especially linear resolution method. Section 7 reviews experimental evidence and competition-based performance validation. Section 8 concludes the paper and outlines future directions.

## 2. Related Works

Automated reasoning has evolved through several major paradigms over the past six decades, from the early symbolic logic programs of the 1950s to modern, learning-guided first-order provers. The central theoretical foundation remains the resolution principle introduced by Robinson (1965), which provided a uniform and complete inference mechanism for first-order logic. Binary resolution, based on complementary literal elimination between two clauses, became the cornerstone of most automated theorem provers, including Vampire (Kovács & Voronkov, 2013), E (Schulz, 2013), Prover9 (McCune, 2025), SPASS (Weidenbach *et al*., 2007). These systems, benchmarked annually in the CADE ATP System Competition (CASC) (CASC 2018-2025; Sutcliffe, 2025), have defined state-of-the-art performance in first-order theorem proving.

### 2.1. From Resolution to Extended Inference Models

Despite its success, binary resolution inherently limits deductive flexibility because each inference involves only two clauses. Over the decades, several extensions have sought to improve reasoning efficiency while retaining logical completeness. Early developments included hyper-resolution (Robinson, 1965), semantic resolution (Chang & Lee, 1973), paramodulation (Robinson & Wos, 1969) and tableaux-based refutation (Fitting 1996), each introducing mechanisms for richer clause interaction. Nevertheless, most of these frameworks remain static and rely on predetermined inference patterns rather than dynamically adaptive multi-clause reasoning.

Later, the superposition calculus (Bachmair & Ganzinger, 1994) emerged as the dominant framework for first-order reasoning with equality, providing the theoretical foundation for saturation-based provers such as E and Vampire. These provers incorporate numerous optimization strategies, such as set-of-support, ordered resolution, subsumption elimination, and clause weighting heuristics control (Riazanov & Voronkov, 2003; Plaisted, 2015; Schulz & Mohrmann, 2016), which collectively improve search efficiency and redundancy, and improve scalability across large problem domains. Yet these methods often trade deductive completeness for computational tractability.

However, all these calculi remain essentially binary in structure. Inference proceeds through pairwise resolution steps, and contradictions are constructed incrementally rather than synergistically. This restriction constrains global clause dependency exploitation and often leads to combinatorial explosion in complex proofs. As demonstrated by the extensive Thousands of Problems for Theorem Provers (TPTP) benchmark library (Sutcliffe, 2017), many challenging problems remain unresolved or inefficiently solved. Consequently, alternative inference paradigms have emerged to model multi-clause reasoning and dynamic inference control beyond traditional resolution frameworks.

**2.2. The Emergence of Contradiction Separation Extension (CSE)**

To overcome the intrinsic limitations of binary reasoning, Xu and colleagues (Xu *et al*., 2018) proposed the **Contradiction Separation Extension (CSE)** framework, a new, dynamic multi-clause reasoning theory for automated deduction. Unlike resolution, which eliminates literals pairwise, CSE introduces the *contradiction separation (CS) rule* which enables deduction steps involving multiple clauses dynamically and synergistically, thereby generalizing binary resolution as a special case. Theoretical results established the *soundness and completeness* of this new CSE framework. CSE provides a more flexible and goal-oriented reasoning structure which transforms logical inference into a dynamic multi-clause synergy process, enabling simultaneous reasoning among interdependent clauses (Chen *et al*., 2018).

**2.3. From CSE Framework to the Extended Triangular Method**

Efficient CSE deduction depends not only on the underlying inference rule but also on how contradictions are constructed and exploited during proof search. The CSE framework has been further elaborated in subsequent studies (Chen *et al*. 2010; Cao *et al*., 2019, 2021; Zhong *et al*., 2020; He *et al*. 2018, 2022), where improved clause-selection strategies and hybrid proof search methods enhanced performance and scalability. In addition, subsequent implementations of CSE-based theorem provers (CASC 2018-2025; Sutcliffe, 2025), such as CSE, CSE_E, CSI_E, and CSI_Enig, demonstrated its potential for higher efficiency and problem-solving capability. Empirical evaluations demonstrated that CSE-based theorem provers can solve a large number of previously unsolved TPTP problems, confirming the practical viability of CSE-based inference. It worth noting that all previous work and CSE based theorem provers used the internal contradiction-construction algorithmic procedures, however, the contradiction construction algorithmic cores have not been documented in formal literature.

Against this backdrop, the Standard Extension (SE) method introduced by Xu *et al*. (2025b), as the first attempt, formalized a specific structured approach in which contradictions are incrementally generated through complementary literal extension, resulting in a standard contradiction structure. SE method provided an important procedural foundation for implementing contradiction-separation-based reasoning. However, it remained limited by the constraint that subsequent deduction clauses must be selected according to existing literal complementarity.

In contrast, the present ETM continues this progression and generalizes that mechanism into a unified triangular geometric framework, allowing broader and more flexible, goal-oriented, and dynamically adaptive selection of clauses and literals during contradiction construction and separation. The "unified" means the ETM not only captures the interaction among three or more clauses but also provides an unified algorithmic model that encompasses earlier contradiction-building strategies, including CSE's foundational rules and CSE's standard extension, within a broader, geometry-inspired triangular framework. Thus, the SE method can be viewed as an early concrete instantiation of the algorithmic principles that are now fully formalized and generalized within the ETM framework.

The ETM introduces a dynamic inference control strategy that generalizes and extends classical resolution by constructing hierarchical triangular relationships among clauses. This approach enables multi-clause coordination, reduces redundancy in the proof search, and supports flexible interaction between deductive and heuristic components. It maintains the theoretical soundness and completeness of the underlying CSE framework while improving the flexibility and efficiency of deduction. ETM's design aligns closely with the needs of dynamic automated reasoning systems, where inference pathways evolve based on clause dependency structures rather than fixed resolution patterns. ETM emphasizes the practical realization of efficient search and reasoning control. It thereby provides a bridge between the theoretical completeness of formal deduction and the scalability required in contemporary AI reasoning systems.

The ETM provides the generalized algorithmic and theoretical foundation underlying the entire family of contradiction-separation-based theorem provers that have competed successfully in the CASC and TPTP benchmark series, including CSE, CSE_E, CSI_E, and CSI_Enig. These results highlight the method's robustness and scalability, demonstrating that multi-clause contradiction construction is not only a theoretical extension but also a practically superior inference strategy. The last four systems demonstrate that the ETM can be integrated with other automated reasoning architectures, enabling hybrid or learning-guided versions of the proposed method.

In summary, chronologically and logically, the SE work represents a prototype algorithmic work, while the ETM is the generalization and consolidation that completes the theoretical–algorithmic bridge.

## 2.4. Integration with Modern Automated Reasoning

Building upon the foundation of **CSE**, further advances have been achieved in automated reasoning through system integration and hybrid inference architectures. The CSE family of automated theorem provers (CASC 2018-2025; Sutcliffe, 2025)—including *CSE, CSE_E, CSI_E,* and *CSI_Enigma*—represents a coherent and progressive evolution of the CSE framework, advancing from dynamic multi-clause reasoning to hybrid, parallel, and learning-guided deduction. Collectively, these systems integrate symbolic synergy, equational reasoning (Kovács & Voronkov, 2013), and neural inference guidance (Jakubuv & Urban, 2020), forming a unified and scalable architecture that advances the state of first-order automated reasoning and demonstrates the algorithmic generality of the CSE theory.

The ETM introduced in this work formalizes and generalizes the core contradiction-construction principle that underlies all CSE-based systems. By enabling flexible clause and literal selection during contradiction construction, the ETM bridges the conceptual gap between classical binary inference, synergized multi-clause reasoning, and modern learning-enhanced deduction. Empirical evidence from competitions such as CASC and TPTP demonstrates that ETM-based systems, such as *CSE_E* (a hybrid of CSE and E's superposition calculus) and *CSI_Enig* (a multi-layer, inverse, and parallel theorem prover that integrates CSE with the ENIGMA learning guidance system), achieve superior performance across diverse logical domains, validating both the robustness and scalability of the approach.

In recent years, machine learning has begun to transform the landscape of automated reasoning. Frameworks such as ENIGMA (Jakubuv & Urban, 2020), DeepHOL (Bansal *et al.*, 2019), and GPT-f (Polu & Sutskever, 2020) employ neural or reinforcement learning models to guide clause selection, premise ranking, and proof search. These approaches represent an important bridge between symbolic deduction and data-driven inference. Within this paradigm, the ETM algorithm naturally complements learning-guided reasoning, as its dynamic clause synergy and contradiction-driven structure can serve as a high-level backbone for integrating symbolic and neural inference. Indeed, hybrid provers such as *CSI_Enig* have already demonstrated the compatibility of ETM's multi-clause inference model with neural-guided clause ranking strategies, establishing a promising path toward adaptive, self-optimizing automated deduction.

The rise of learning-guided reasoning—notably the ENIGMA framework (Jakubuv & Urban, 2020), which applies gradient-boosted and neural models to predict useful clauses—has yielded substantial performance improvements in large-theory reasoning and influenced new systems such as DeepHOL (Bansal *et al.*, 2019) and neural premise-selection modules in Lean and Coq. This evolution from static heuristics to data-driven inference marks a major shift toward AI-assisted theorem proving, uniting symbolic logic and modern machine learning.

In conclusion, the ETM represents a significant step toward unified, adaptive, and learning-compatible automated reasoning, bridging classical logical theory with contemporary AI-driven inference paradigms. It builds upon the established strengths of resolution and superposition calculi while introducing a flexible geometric model for contradiction construction and multi-clause synergy, advancing the broader goal of scalable, dynamic reasoning in Artificial Intelligence.

## 3. Preliminaries

This section provides the main concepts and results about soundness and completeness of the contradiction separation based dynamic automated deduction (CSE) theory [Xu *et al*. 2018]. We need the following preliminaries first.

In propositional logic, we consider propositional formula in *conjunctive normal forms* (CNF) which are defined as follows:

- A *literal* is either a propositional logic variable $p$ or its negation $\sim p$.
- Two literals are said to be *complements* or *a complementary pair* if one is the negation of the other (e.g., $\sim p$ is taken to be the complement to $p$).
- A *clause C*, an expression formed from a finite collection of literals, is a disjunction of literals usually written as $C = p_1 \vee p_2 \vee \cdots \vee p_k$, where $p_i$ is a literal. A clause can be empty (defined from an empty set of literals), denoted by $\emptyset$. The truth evaluation of an empty clause is always *false*.
- A *formula* is a conjunction of clauses. A formula $S = C_1 \wedge \cdots \wedge C_m$ in CNF is usually regarded a set of clauses, written as $S = \{C_1, C_2, \ldots, C_m\}$.
- A formula is said to be satisfiable if it can be made TRUE by assigning appropriate logical values (i.e. TRUE, FALSE) to its variables.

In the first-order logic:

- A *literal* is either an atom or a negated atom, where an *atom* is an $n$-ary predicate (denoted $P$ or $Q$) applied to $n$ terms.
- A *term* is either a constant (denoted $a$ or $b$), a variable (denoted $x$, $y$, $v$ or $z$) or an $n$-ary function

(denoted *f* or *g*) applied to *n* terms.
- A *clause* is simply a disjunction of literals where all variables are universally quantified.

**Remark 3.1** In propositional logic and first-order logic, if there are the same literals in a clause, we regard these same literals to be one literal.

- *Substitutions* (denoted by $\sigma$, possibly superscripted) is a mapping from variables to terms. Considering a clause *C*, we write $C^\sigma$ to denote the result of substituting each assigned variable with the assigned term in *C*. The empty (i.e. identity) substitution is denoted $\emptyset$. If none of the terms in a substitution contains a variable, i.e., all the terms in the substitution are *ground terms*, we have a so-called *ground substitution*. If $\sigma$ is a (ground) substitution, then $C^\sigma$ is called an (*ground*) *instance* of *C*.

- A substitution $\sigma$ is a *unifier* of logical expressions $e_1, \ldots, e_n$ if and only if $\sigma(e_1)=\ldots=\sigma(e_n)$ where "=" denotes syntactic identity. We refer the reader to, for instance, [6, 13, 17], for more details about logical notations and resolution concept.

**Definition 3.1** (Xu *et al.* 2018) **(Contradiction)** Assume a clause set $S=\{C_1, C_2, \ldots, C_m\}$ in propositional logic. If for any $(p_1, \ldots, p_m) \in \prod_{i=1}^{m} C_i$, $p_i$ (*i*=1,…, *m*) is a literal, there exist some complementary pair of literals among $p_1,\ldots,p_m$, then $S=\bigwedge_{i=1}^{m} C_i$ is called a *standard contradiction*. If $\bigwedge_{i=1}^{m} C_i$ is unsatisfiable, then $S=\bigwedge_{i=1}^{m} C_i$ is called a *quasi-contradiction*, where $C_i$ is also regarded as a set of literals (*i*=1, …, *m*).

**Lemma 3.1** (Xu *et al.* 2018) Assume a clause set $S=\{C_1, C_2, \ldots, C_m\}$ in propositional logic, then *S* is a standard contradiction if and only if *S* is a quasi-contradiction. Therefore, in propositional logic, both standard contradiction and quasi-contradiction are shortly called *contradiction*.

**Definition 3.2** (Xu *et al.* 2018) (**Contradiction Separation Clause in Propositional Logic**) Assume a clause set $S=\{C_1, C_2, \ldots, C_m\}$ in propositional logic. A new clause is called a *contradiction separation clause* (CSC) of $C_1, C_2, \ldots, C_m$, denoted as $\mathbb{C}_m(C_1, C_2, \ldots, C_m)$, if the following conditions hold:

(1) For any $C_i$ (*i*=1,…, *m*), $C_i$ can be partitioned into two sub-clauses $C_i^-$ and $C_i^+$ such that $C_i = C_i^- \vee C_i^+$, where no same literal exists in $C_i^-$ and $C_i^+$; $C_i^+$ can be an empty clause itself, but $C_i^-$ cannot be an empty clause; moreover,

(2) $\bigwedge_{i=1}^{m} C_i^-$ is unsatisfiable; and

(3) $\bigvee_{i=1}^{m} C_i^+ = \mathbb{C}_m (C_1, C_2, \ldots, C_m)$.

The inference rule that produces a new clause $\mathbb{C}_m(C_1, C_2, \ldots, C_m)$ is called *a contradiction separation rule*, in short, a CS rule.

It can be seen from Def. 3.2 that contradiction separation rule actually splits each selected clause into two parts, denoted with positive and negative signs in Def 3.2, in such a form that the conjunction of the parts with negative sign form a contradiction, and the disjunction of the remaining literals of these clauses, i.e., the parts with positive sign, is the intended output of the rule.

**Definition 3.3** (Xu *et al*. 2018) Suppose a clause set $S = \{C_1, C_2, \ldots, C_m\}$ in propositional logic. $\Phi_1$, $\Phi_2, \ldots, \Phi_t$ is called a *contradiction separation based dynamic deduction sequence* (*or CS based dynamic deduction sequence) from S to a clause $\Phi_t$*, denoted as $D$, if for $i \in \{1, 2, \cdots, t\}$,

(1) $\Phi_i \in S$, or

(2) there exist $r_1, r_2, \ldots, r_{k_i} < i$, $\Phi_i = \mathbb{C}_{k_i}(\Phi_{r_1}, \Phi_{r_1}, \ldots, \Phi_{r_{k_i}})$.

**Theorem 3.1** (Xu *et al*. 2018) (**Soundness Theorem of the CS-Based Deduction in Propositional Logic**) Suppose a clause set $S = \{C_1, C_2, \ldots, C_m\}$ in propositional logic. $\Phi_1, \Phi_2, \ldots, \Phi_t$ is a CS based deduction sequence from $S$ to a clause $\Phi_t$. If $\Phi_t$ is an empty clause, then $S$ is unsatisfiable.

**Theorem 3.2** (Xu *et al*. 2018) (**Completeness Theorem of the CS-Based Deduction in Propositional Logic**) Suppose a clause set $S = \{C_1, C_2, \ldots, C_m\}$ in propositional logic. If $S$ is unsatisfiable, then there exists a CS based deduction sequence from $S$ to an empty clause.

**Definition 3.4** (Xu *et al*. 2018) (**Contradiction in First-Order Logic**) Suppose a clause set $S = \{C_1, C_2, \ldots, C_m\}$ in first-order logic. A new clause is called a *standard contradiction separation clause* (S-CSC) of $C_1, C_2, \ldots, C_m$, denoted as $\mathbb{C}_m^{s\sigma}(C_1, C_2, \ldots, C_m)$ (here "$s$" means "standard", and $\sigma = \bigcup_{i=1}^{m} \sigma_i$, $\sigma_i$ is a substitution to $C_i$, $i=1,\ldots, m$), if the following conditions hold:

(1) there does not exist the same variables among $C_1, C_2, \ldots, C_m$ (if there exist the same variables, there will be a rename substitution to make them different);

(2) For any $C_i$, $i=1, 2, \ldots, m$, a substitution $\sigma_i$ can be applied to $C_i$ ($\sigma_i$ could be an empty substitution) and the same literals merged after substitution, denoted as $C_i^{\sigma_i}$; in addition, $C_i^{\sigma_i}$ can be partitioned into two sub-clauses $C_i^{\sigma_i^-}$ and $C_i^{\sigma_i^+}$ such that

i) $C_i^{\sigma_i} = C_i^{\sigma_i^-} \vee C_i^{\sigma_i^+}$, where $C_i^{\sigma_i^-}$ and $C_i^{\sigma_i^+}$ do not share the same literal, $C_i^{\sigma_i^+}$ can be an empty clause, $C_i^{\sigma_i^-}$ cannot be an empty clause; moreover,

ii) $\bigwedge_{i=1}^{m} C_i^{\sigma_i^-}$ or $S$ is *a standard contradiction*, that is for any $(x_1, \ldots, x_m) \in \prod_{i=1}^{m} C_i^{\sigma_i^-}$, there exists at least one complementary pair among $\{x_1, \ldots, x_m\}$;

iii) $\bigvee_{i=1}^{m} C_i^{\sigma_i^+} = \mathbb{C}_m^{s\sigma}(C_1, C_2, \ldots, C_m)$.

The inference rule that produces a new clause $\mathbb{C}_m^{s\sigma}(C_1, C_2, \ldots, C_m)$ is called *a standard contradiction separation rule* in first-order logic, in short, a S-CS rule.

**Definition 3.5** (Xu *et al*. 2018) Suppose a clause set $S = \{C_1, C_2, \ldots, C_m\}$ in first-order logic. A new clause is called a *quasi-contradiction separation clause* (Q-CSC) of $C_1, C_2, \ldots, C_m$, denoted as $\mathbb{C}_m^{q\sigma}(C_1, C_2, \ldots, C_m)$ (here "$q$" means "quasi", and $\sigma = \bigcup_{i=1}^{m} \sigma_i$, $\sigma_i$ is a substitution to $C_i$, $i=1,\ldots, m$), if the following

conditions hold:

(1) there does not exist the same variables among $C_1, C_2, \ldots, C_m$ (if there exist the same variables, a rename substitution will be applied to make them different);

(2) For any $C_i$, $i=1, 2,\ldots, m$, a substitution $\sigma_i$ can be applied to $C_i$ ($\sigma_i$ could be an empty substitution) and the same literals merged after substitution, denoted as $C_i^{\sigma_i}$; in addition, $C_i^{\sigma_i}$ can be partitioned into two sub-clauses $C_i^{\sigma_i^-}$ and $C_i^{\sigma_i^+}$ such that

i) $C_i^{\sigma_i} = C_i^{\sigma_i^-} \vee C_i^{\sigma_i^+}$, where $C_i^{\sigma_i^-}$ and $C_i^{\sigma_i^+}$ do not share the same literal, $C_i^{\sigma_i^+}$ can be an empty clause, $C_i^{\sigma_i^-}$ cannot be an empty clause; moreover,

ii) $\bigwedge_{i=1}^{m} C_i^{\sigma_i^-}$ is unsatisfiable, $\bigwedge_{i=1}^{m} C_i^{\sigma_i^-}$ or $S$ is called *a quasi-contradiction*;

iii) $\bigvee_{i=1}^{m} C_i^{\sigma_i^+} = \mathbb{C}_m^{q\sigma}(C_1, C_2,\ldots, C_m)$.

The inference rule that produces a new clause $\mathbb{C}_m^{q\sigma}(C_1, C_2, \ldots, C_m)$ is called *a quasi-contradiction separation rule* in first-order logic, in short, a Q-CS rule.

**Lemma 3.2** (Xu *et al*. 2018) Suppose a clause set $S = \{C_1, C_2, \ldots, C_m\}$ in first-order logic. If $C_1 \wedge C_2 \wedge \ldots \wedge C_m$ is a standard contradiction, then $S$ is a quasi-contradiction (i.e., $S$ is unsatisfiable). On the other hand, if $S = \{C_1, C_2, \ldots, C_m\}$ is unsatisfiable, it does not mean that $S$ is a standard contradiction.

**Corollary 3.1** (Xu *et al*. 2018) (**Invariance of Standard Contradiction in terms of Variable Substitution**) Suppose $S = \{C_1, C_2, \ldots, C_m\}$, where $C_1, C_2,\ldots, C_m$ are clauses in the first order logic. If $C_1 \wedge C_2 \wedge \ldots \wedge C_m$ is a standard contradiction, then for any variable substitution for $\sigma$ of $S$, $(C_1 \wedge C_2 \wedge \ldots \wedge C_m)^\sigma$ is also a standard contradiction.

**Definition 3.6** (Xu *et al*. 2018) Suppose a clause set $S = \{C_1, C_2, \ldots, C_m\}$ in first-order logic. $\Phi_1, \Phi_2,\ldots, \Phi_t$ is called *a standard contradiction separation based deduction sequence* (or a ***S-CS** based deduction sequence* from $S$ to a clause $\Phi_t$, denoted as $D^s$, if for $i \in \{1,2,\cdots,t\}$,

(1) $\Phi_i \in S$, or

(2) there exist $r_1, r_2, \ldots, r_{k_i} < i$, $\Phi_i = \mathbb{C}_m^{s\sigma}(\Phi_{r_1}, \Phi_{r_1}, \ldots, \Phi_{r_{k_i}})$.

**Theorem 3.3** (Xu *et al*. 2018) (**Soundness Theorem of the S-CS Based Deduction in First-Order Logic**) Suppose a clause set $S = \{C_1, C_2, \ldots, C_m\}$ in first-order logic. $\Phi_1, \Phi_2, \ldots, \Phi_t$ is a S-CS based deduction from $S$ to a clause $\Phi_t$. If $\Phi_t$ is an empty clause, then $S$ is unsatisfiable.

**Theorem 3.4** (Xu *et al*. 2018) (**Completeness of the S-CS Based Deduction in First-Order Logic**) Suppose a clause set $S = \{C_1, C_2, \ldots, C_m\}$ in first-order logic. If $S$ is unsatisfiable, then there exists an S-CS based deduction from $S$ to an empty clause.

Table 3.1 below summaries the features of CSE in a breaking down structure.

**Table 3.1** Features of CSE in a breaking down structure

| Component | Meaning |
|---|---|
| **Contradiction Separation (CS)** | The foundational inference principle — instead of resolving only *two complementary literals* (as in binary resolution), CS identifies and separates *multi-clause sets* that collectively form a contradiction. It treats contradictions as *cooperative clause structures*, not just binary pairs. |
| **Dynamic** | The inference process is not static; clause combinations are chosen adaptively during proof search, allowing the system to dynamically reconfigure which clauses interact. |
| **Multi-Clause Synergized** | Multiple clauses work *synergistically* in one deduction step, creating a more expressive and potentially shorter reasoning path than traditional pairwise resolution. |
| **Automated Deduction** | The entire framework is an automated theorem-proving mechanism for first-order logic (FOL) and propositional logic. |

# 4. Extended Triangular Based Contradiction Separation Automated Deduction Method in Propositional Logic

The key idea of contradiction separation based automated deduction method is to construct contradiction from given clause set, which usually contains more literals from some clauses than just a complementary pair of literals as in binary resolution method, and separate it to allow the remaining literals to generate a new clause by taking their disjunction, i.e., CSC as defined in Def. 3.2 and Def. 3.4. The generated CSC is then added to the original clause set, and the process of constructing contradiction and generating CSC will continue until an empty CSC is generated or certain stopping conditions are achieved. The conclusion that the original clause set is unsatisfiable can then be drawn. The crucial part in the above process is how to construct contradiction during each step.

The ETC based contradiction separation automated deduction method, *Extended Triangular Method* (ETM) in short, proposed in this paper provides an effective contradiction construction method. It is more flexible, comparing with the Standard Extension (SE) method in [Xu *et al*. 2025], by removing the requirement that the selection of subsequent deduction clause must rely on a complementary pair. The constructed contradiction by applying ETM is called as *Extended Triangular Contradiction* (ETC in short). This section introduces the concept and properties of ETC firstly, and then discusses the ETM method in propositional logic along with some strategies for constructing ETC. ETM in first-order logic will be discussed in the next section.

## 4.1 Extended Triangular Contradiction (ETC)

Before giving the concept, the idea of the ETC is intuitively depicted as in Fig. 4.1 and Fig. 4.2.

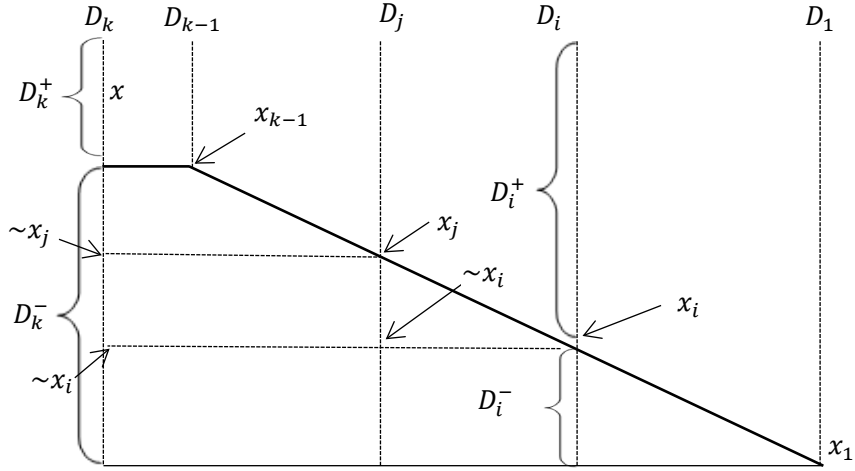

Fig. 4.1 Illustration figure of ETC

| $D_k$ | $D_{k-1}$ |  |  |  | $D_{i+1}$ | $D_i$ | $D_{i-1}$ |  |  | $D_3$ | $D_2$ | $D_1$ |
|---|---|---|---|---|---|---|---|---|---|---|---|---|
|  |  |  |  |  |  |  |  |  |  |  |  |  |
|  |  |  |  |  |  |  |  |  |  |  |  |  |
|  | $x_{k-1}$ |  |  |  |  |  |  |  |  |  |  |  |
|  |  | $x_{k-2}$ |  |  |  | $y_i$ |  |  |  |  |  |  |
|  |  |  |  |  |  |  |  |  |  |  |  |  |
|  |  |  |  |  |  |  |  |  |  |  |  |  |
|  |  |  |  |  | $x_{i+1}$ |  | $y_{i-1}$ |  |  |  |  |  |
| $\sim x_i$ |  |  |  |  | $\sim x_i$ | $x_i$ |  |  |  |  |  |  |
|  |  |  | $\sim x_{i-1}$ |  |  |  | $x_{i-1}$ |  |  |  | $y_2$ |  |
|  |  |  |  |  |  |  |  |  |  |  |  |  |
|  |  |  |  |  |  |  |  |  |  |  |  |  |
|  |  |  |  |  |  |  |  |  |  | $x_3$ |  |  |
|  |  | $\sim x_2$ |  |  |  |  | $\sim x_2$ |  |  |  | $x_2$ |  |
|  |  |  | $\sim x_1$ |  |  |  | $\sim x_1$ |  |  |  |  | $x_1$ |

Fig. 4.2 Tabular form of ETC

Given clause set $S = \{C_1, C_2, \ldots, C_m\}$ ($m \geq 2$) in propositional logic, the form of ETC is illustrated in Fig. 4.1 and Fig. 4.2, where each column denotes the literals in clause $D_i$, $i=1, 2, \ldots, k$ ($k \geq 2$), selected from $S$ ($S$ may include generated CSC in the real deduction process), and each row contains the selected literal $x_i$ itself and possible complementary literals $\sim x_i$ to $x_i \in D_i$, $i=1, 2, \ldots, k$-1 ($k \geq 2$).

The selection of $D_i$ ($i=1, \ldots, k$-1) from $S$ relies mainly on the selection of certain literal in it, i.e., $x_1, \ldots, x_{k-1}$ in Fig. 4.1 or Fig. 4.2. These literals from different clauses and the possible literal $\sim x_{k-1}$ from the last clause $D_k$ in this deduction step form a slant line as shown in the illustration figure, which actually serves as the boundary splitting the contradiction and CSC parts, and so called the main boundary line. The complementary literal, if there is, in $D_i$ ($i=1, \ldots, k$), say $\sim x_j$ ($1 \leq j \leq i-1$), to the literals $x_1, \ldots, x_{i-1}$ on the main boundary line is put at the same row where $x_j$ locates.

It can be seen from Fig. 4.1 and Fig. 4.2 that each clause $D_i$ is divided by the main boundary line into two parts, i.e., $D_i = D_i^+ \cup D_i^-$ ($i=1, ..., k$), where $D_i^+$ is the set of literals above the main boundary line, and $D_i^-$ contains the literals below and on the main boundary line, as shown in Fig. 4.1, i.e., $D_i^- \subseteq \{\sim x_1, \cdots, \sim x_{i-1}, x_i\}$ for $i=1, 2, ..., k$-1 ($k \geq 2$) and $\emptyset \neq D_k^- \subseteq \{\sim x_1, \cdots, \sim x_{k-1}\}$. By doing this, each clause $D_i$ ($i=1, ..., k$) is actually divided into two sub-clauses $D_i^+$ and $D_i^-$, where both the literals in $D_i^+$ and $D_i^-$ take disjunction form. It can be seen that, $\bigwedge_{i=1}^{k} D_i^-$ is actually a contradiction, and called an *Extended Triangular Contradiction* (ETC), where the formal definition is provided in Def. 4.1.

**Definition 4.1** Assume a clause set $S = \{C_1, C_2, ..., C_m\}$ ($m \geq 2$) in propositional logic, where $D_1, D_2, ..., D_k$ ($k \geq 2$) are the clauses selected for contradiction construction in one deduction step. Each $D_i$ can be partitioned into two sub-clauses $D_i^-$ and $D_i^+$ sharing no same literal, such that $D_i = D_i^- \vee D_i^+$, $i=1, ..., k$. $\bigwedge_{i=1}^{k} D_i^-$ is called an ETC, if the following conditions hold.

(1) For $D_i = D_i^- \vee D_i^+$, $i=1, ..., k$, $D_i^+$ can be an empty sub-clause, but $D_i^-$ cannot be empty, where there is no duplicated literal in $D_i^-$; moreover,

(2) There are literals $x_j \in D_j$ ($j=1, ..., k$-1), such that $\emptyset \neq D_k^- \subseteq \{\sim x_j | j = 1, \cdots, k-1\}$, where $D_i^- = \{x_i\} \cup D_i^0$, $D_i^0 \subseteq \{\sim x_t | t = 1, ..., i - 1\}$ ($i=2,...,k$-1), $D_1^- = \{x_1\}$.

As stated previously, $\{x_1, \cdots, x_{k-1}\}$ and the possible literal $\sim x_{k-1}$ from the last clause $D_k$ in this deduction step is called the main boundary line.

The clause with less literals is usually selected first during the contradiction construction process, which makes the shape below (including) the main boundary line, i.e., the contradiction part, take the triangle like shape. Therefore, this kind of CS-based automated deduction method is named as Triangular method, including the Standard Triangular method (Xu *et al*. 2025b) and the Extended Triangular method proposed in this paper. Considering the clause and literal selection, the clause with small $|D_i^+|$ is usually preferred, where $|D|$ means the number of literals in $D$, while the concrete selection method in Extended Triangular method will be discussed in the following sub-sections.

**Remark 3.1** (1) The clauses used for constructing ETC can be used repeatedly, and the literals in the main boundary line can also be used repeatedly.

(2) There is no complementary pair of literals in the main boundary line.

**Remark 3.2** Condition in Def. 4.1 that there is no duplicated literal in $D_i^-$ means that,

(1) As it is allowed that the same literal, for example $x$, can appear in the main boundary line repeatedly, and if there is $\sim x$ in $D_i$, $\sim x$ is allowed to appear only once.

(2) If the repeated literal is $\sim x_{k-1}$ from the last clause $D_k$, it is preferred to put the literal $\sim x_{k-1}$ at the same line with $x_{k-1}$, so that a "flat top" will appear at the leftmost side of the ETC as shown in Fig. 4.1. This kind of "flat top" is a visible sign for determining the contradiction.

**Lemma 4.1** [Xu *et al*. 2025a] Suppose that $S = \{D_1, \cdots, D_k\}$ ($k \geq 2$) is a set of clauses in propositional logic. If $D_1 = \{x_1\}$, $D_i = \{x_i\} \cup \{\sim x_t | t = 1, ..., i - 1\}$ ($i=2,...,k$-1), $D_k = \{\sim x_j | j = 1, \cdots, k - 1\}$, then the following statements hold.

(1) $\bigwedge_{i=1}^{k} D_i$ is a standard contradiction.

(2) Taking $D_i$ $(i = 1, \cdots, k)$ as a set of literals, then the corresponding clause set of each subset $D^*$ of their Cartesian product $\prod_{i=1}^{k} D_i$ is a standard contradiction.

$\bigwedge_{i=1}^{k} D_i$ is called as full standard triangular contradiction, and $D^*$ is a normal contradiction.

**Theorem 4.1** The ETC defined in Def 4.1, i.e., $\bigwedge_{i=1}^{k} D_i^-$, is a contradiction.

**Proof**. From conditions (1) and (2) in Def. 4.1, there is no different atom in $D_k^-$ to the atoms appearing in $\bigwedge_{i=1}^{k-1} D_i^-$. In other word, the number of atoms in $\bigwedge_{i=1}^{k} D_i^-$ is not greater than $k$-1, and each sub-clause $D_i^-$ is not empty. It means that, $\prod_{i=1}^{k} D_i^-$ is an non-empty subset of the Cartesian product corresponding to the full standard triangular contradiction $\{\sim x_{k-1}, \sim x_{k-2}, \cdots, \sim x_1\} \times \prod_{i=1}^{k-1}\{x_i, \sim x_{i-1}, \cdots, \sim x_1\}$. Therefore, it can be concluded based on Lemma 4.1 that the ETC defined in Def. 4.1, i.e., $\bigwedge_{i=1}^{k} D_i^-$, is a contradiction.

The disjunction of the literals above the ETC, excluding the main boundary line, i.e., $\bigvee_{i=1}^{k} D_i^+$, is the corresponding contradiction separation clause.

**Remark 4.3** Theorem 4.1 shows that the requirement of a "flat top" appearing at the leftmost side of the ETC is not necessary. This is actually the difference between ETC and Standard Triangle.

On the other hand, we can fill some literals to the ETC defined in Def. 4.1, to make it a full standard triangular contradiction, i.e., $D_k^- = \{\sim x_j | j = 1, \cdots, k-1\}$, $D_i^0 = \{\sim x_j | j = 1, \ldots, i-1\}$ ($i$=2,..., $k$-1), while the corresponding CSC will remain the same as before this kind of filling. That is, the following proposition holds.

**Corollary 4.1** ETC can be expanded to a full standard triangular contradiction.

**Corollary 4.2** If $D_k^+ \neq \emptyset$, then, for any $x \in D_k^+$, $x_1$, ..., $x_{k-1}$, $x$ is a set satisfiable instances for the set of clauses $D_1, D_2, \ldots, D_k$ used for constructing the ETC.

Table 4.1 The constructed ETC in Example 4.1

| $C_3$ | $C_4$ | $C_2$ | $C_1$ |
|---|---|---|---|
|  |  |  |  |
| $p_3$ | $\sim p_3$ |  |  |
| $\sim p_4$ |  | $p_4$ |  |
|  | $\sim p_1$ | $\sim p_1$ | $p_1$ |

**Example 4.1** Suppose that $S=\{C_1, C_2, C_3, C_4\}$ is a clause set in propositional logic, where $C_1 = p_1$, $C_2 = \sim p_1 \vee p_4$, $C_3 = p_3 \vee \sim p_4$, $C_4 = \sim p_1 \vee \sim p_3$, with $p_1, p_2, p_3, p_4$ being propositional variables. We can then construct the ETC with $C_1^- = p_1$, $C_1^+ = \emptyset$, $C_2^- = \sim p_1 \vee p_4$, $C_2^+ = \emptyset$, $C_3^- = p_3 \vee \sim p_4$, $C_3^+ = \emptyset$, $C_4^- = \sim p_1 \vee \sim p_3$, and $C_4^+ = \emptyset$, as shown in Table 4.1. It can be seen that the corresponding CSC is $\bigvee_{i=1}^{4} C_i^+ = \emptyset$, which means that the clause set $S$ is unsatisfiable.

## 4.2 Different Forms of ETC

As a special type of contradiction, the ETC has some different representation forms due to its flexibility on construction. We list some popular ones in this subsection.

**(1) Flat-top ETC**

The form with a flat top at the leftmost side of ETC is the commonest one, and also the easiest form to determine the unsatisfiability of the given clause set. It is characterized by the existence of literal $\sim x_{k-1}$ in the last column, i.e., the last clause $D_k$, which locates at the same row with $x_{k-1}$ in clause $D_{k-1}$. There is a visible "flat top" at the leftmost two columns for this type of ETC (Fig. 4.1).

**(2) Standard Triangle contradiction**

ETC is an extension of Standard Triangle, and Standard Triangle is a special case of ETC. When the complementary literal $\sim x_i$ does exist in $D_{i+1}$ for each literal $x_i$ ($i$=2, …, $k$-1) in the main boundary line, the ETC will turn into a Standard Triangle.

**(3) ETC with stair-like main boundary line**

For the clause set $S$={$D_1$, …, $D_k$} in propositional logic, suppose that the constructed ETC is as shown in Fig. 4.3, where $D_{i_0}^+ = \emptyset, D_{i_0}^-(\neq \emptyset) \subseteq \{\sim x_{i_0-1}, \cdots, \sim x_1\}$, i.e., there is a flat "stair" at the place of clause $D_{i_0}$ in the main boundary line. This kind of flat "stair" is formed because there is no main boundary line literal in $D_{i_0}$. This kind of contradiction is a transformation of ETC, which does not affect its unsatisfiability, and so-called ETC with stair-like main boundary line. It can be easily seen that this kind of "stair" may appear several times, even one followed by another.

For ETC with stair-like main boundary line, $D_{i_0}^+ = \emptyset$, $D_{i_0}^-(\neq \emptyset) \subseteq \{\sim x_{i_0-1}, \cdots, \sim x_1\}$, which means that there is no main boundary line literal $x_{i_0}$ in $D_{i_0}$, and therefore there is no complementary literal $\sim x_{i_0}$ in $D_k$,…, $D_{i_0+1}$ corresponding to $x_{i_0}$. The corresponding contradiction is $\wedge_{i=1}^{i_0-1} D_i^- \wedge D_{i_0}^- \wedge_{i=i_0+1}^{k} D_i^-$.

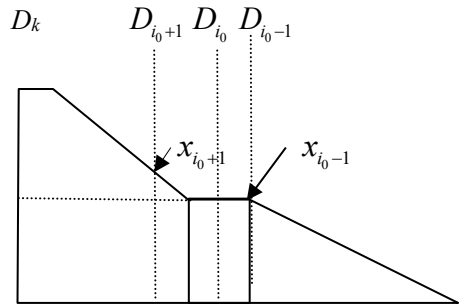

Fig. 4.3 ETC with stair-like main boundary line (△1)

The flat "stair" in Fig. 4.3 can be moved to the leftmost side, and the obtained contradiction is as shown in Fig. 4.4, which is also a kind of transformation of ETC. The corresponding contradiction is $D_{i_0}^- \wedge \wedge_{\substack{i=1 \\ i \neq i_0}}^{k} D_i^-$.

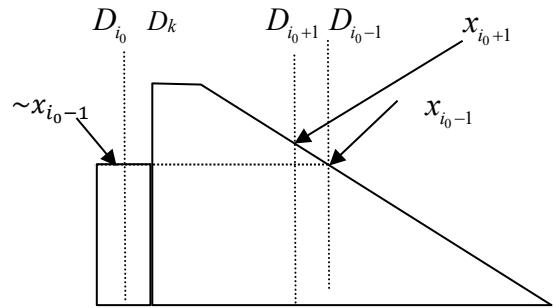

Fig. 4.4 ETC with stair-like main boundary line after transformation (△2)

**Remark 4.4** i) For △2 shown in Fig. 4.4, the obtained CSC is exactly the same as that obtained by separating the contradiction △1 from $S$. It means that the deduction result is not affected by moving $D_{i_0}$ to the left side.

ii) △2 takes the same essential property, i.e., unsatisfiability, as △1, and so this kind of transformation is called as *deduction reserving transformation*.

iii) This transformation is actually moving all the clauses containing no literal in the main boundary line to the left side of the constructed ETC.

**Remark 4.5** For ETC with stair-like main boundary line where $D_{i_0}^+ = \emptyset$, $D_{i_0}^-(\neq \emptyset) \subseteq \{\sim x_{i_0-1}, \cdots, \sim x_1\}$,

i) The sub-clauses $D_{i_0}^-$, $D_{i_0-1}^-$, …, $D_1^-$ form a contradiction already, where $D_{i_0}^-$ is necessary.

ii) If $i_0 < k$, i.e., $D_{i_0}$ is not the last clause in this deduction step, then $D_{i_0}$ is a redundant clause during this deduction step, because it does not contain literal $x_{i_0}$ and therefore there is no link (complementary literals) between $D_{i_0}$ and other clauses. This kind of clause $D_{i_0}$ can be removed from the deduction process without affecting the other clauses and the deduction result.

iii) If there is no complementary literal to $x_{i_0}$ in the constructed ETC where there is no other literal in the same row with $x_{i_0}$, i.e., $\sim x_{i_0} \notin D_i$ for $i = i_0+1,…, k$, then $D_{i_0}$ is a redundant clause during this deduction step, and we can remove $D_{i_0}$ from this ETC without affecting the deduction result.

(4) ETC with only main boundary line

As a special case, ETC can be the case that there are literals only in the main boundary line, i.e., the bold line in Fig. 4.1, while no literal in other places. This type of ETC has the least number of literals under the same number of clauses with the only complementary pair of literals $x_{k-1}$ and $\sim x_{k-1}$.

**4.3 ETC construction Method in Propositional Logic**

We provide a general framework for constructing the ETC in propositional logic.

Given a clause set $S = \{C_1, C_2, …, C_m\}$ ($m \geq 2$) in propositional logic, we can select clauses $D_1, D_2, …, D_k$ ($k \geq 2$) from $S$ and the corresponding main boundary line literals successively based on the following steps, to construct the ETC.

**Step 1. Selection of the first clause $D_1$ and the main boundary line literal $x_1$**

Select clause $D_1$ (usually with least number of literals) from the clause set $S$ and the corresponding literal $x_1 \in D_1$, denote $D_1^- = \{x_1\} \neq \emptyset$ and $D_1^+ = D_1 - D_1^-$, where the literals in $D_1^-$ and $D_1^+$ are in disjunction form, i.e., $D_1^-$ and $D_1^+$ form two sub-clauses, where $D_1^+$ might be empty.

…

**Step $i$. Selection of clause $D_i$ ($2 \leq i \leq k-1$) and the main boundary line literal $x_i$**

Select clause $D_i$ from the clause set $S$ and the corresponding literal $x_i \in D_i$, and put the literals in $D_i$, which are complementary to the literals $x_j \in D_j$ ($j \in \{1, 2, \cdots, i-1\}$) selected in the previous $i$-1 steps, on the same row where $x_j$ appears. Denote $D_i^- = \{x_i\} \cup D_i^0 \neq \emptyset$, $D_i^0 \subseteq \{\sim x_j | j = 1, …, i-1\}$, $D_i^+ = D_i - D_i^-$, where the literals in $D_i^-$ and $D_i^+$ are in disjunction form, i.e., $D_1^-$ and $D_1^+$ form two sub-clauses. $D_i^-$ is the set of literals of $D_i$ within ETC, including the main boundary line, and $D_i^0$ is the set of leftover literals in $D_i^-$ after removing $\{x_i\}$.

…

**Step $k$. Selection of clause $D_k$ (the construction process stops at this step)**

Select clause $D_k$ from the clause set $S$ and let $\emptyset \neq D_k^- \subseteq \{\sim x_j | j = 1, ..., k-1\}$, $D_k^+ = D_k - D_k^-$, where the literals in $D_k^-$ and $D_k^+$ are in disjunction form, i.e., $D_k^-$ and $D_k^+$ form two sub-clauses.

This round of ETC stops at Step $k$, where the stop conditions can be found in Remark 4.6 (2), and the obtained CSC is $R_s = \bigvee_{i=1}^{k} D_i^+$. Actually, we have the following propositions.

**Lemma 4.2** The ETC constructed from the clause set $S$ may include all the clauses in $S$.

**Proof.** Given a clause set $S=\{C_1, C_2, ..., C_m\}$ ($m \geq 2$) in propositional logic, for any clause $D \in S$,

(1) If $D$ is $D_1$, i.e., the first selected clause with a selected literal $x \in D_1$, in ETC construction, then it can be easily seen that $D$ participates the construction of ETC.

(2) Suppose that the clauses $D_1, \cdots D_i$ in $S$ participate the construction of ETC successively according to the requirements of ETC, and the sequence of main boundary line literals $z_1, \cdots, z_t$ are selected from the corresponding clause $D_j \in \{D_1, \cdots, D_i\}$ respectively.

If $S - \{D_1, \cdots, D_i\} = \emptyset$, then all the clauses in $S$ have been included in the constructed ETC. If $S - \{D_1, \cdots, D_i\} \neq \emptyset$, then for any $D \in S - \{D_1, \cdots, D_i\}$,

i) If there is literal $z_{t+1} \in D$, $z_{t+1} \notin \{\sim z_1, \cdots, \sim z_t\}$, the we can select $z_{t+1}$ as the main boundary line literal, let $D$ be $D_{i+1}$, and continue the ETC construction process. If $D$ is the last clause left in $S$, and $D^- = \emptyset$, then the literals in $D$ are all pure literals, or their complementary literals belong to $D_1^+, \cdots, D_i^+$, and then $D$ is a redundant clause in $S$ and can be removed.

ii) If there is no literal $z_{t+1} \in D$, $z_{t+1} \notin \{\sim z_1, \cdots, \sim z_t\}$, i.e., $D \subseteq \{\sim z_1, \cdots, \sim z_t\}$, then it means that $D^- \subseteq \{\sim z_1, \cdots, \sim z_t\}$ and $D^+ = \emptyset$. We can still let $D$ be $D_{i+1}$, and continue the ETC construction. The obtained ETC will take the form of stair-like main boundary line, and this kind of clause can be moved to the leftmost side of the ETC as described in subsection 4.2.

Based on the above process, all the clauses in $S$ can be put into an ETC. This completes the proof.

**Remark 4.6** (1) During the ETC construction process, there is actually no special restriction on clause $D_i$ and the corresponding literal $x_i$ ($i=1, 2, ..., k$-1). On the other hand, some specific methods and strategies are usually provided in order to improve the efficiency and ability of the deduction system based on ETC, which will be discussed in detail in Subsection 4.6.

(2) The selection of the last clause $D_k$ of this round ETC construction is a bit different to that of previous clauses. It is required that there must be literal in $D_k$ complementary to some literal in the main boundary line, which is to guarantee that the constructed ETC is a contradiction.

(3) From Lemma 4.2, we know that all the clauses in $S$ can be included in the ETC constructed based on $S$. In order to improve the efficiency of ETC based deduction system, the stop condition can be adopted as $D_k^- \neq \emptyset$ and $D_k^+ = \emptyset$, and can also be,

i) $\forall x \in D_k^+$, let $S_k(x) = \{D \in S | \exists y \in D, y = \sim x\}$. If $S_k(x) = \emptyset$, then the construction of ETC stops at step *k*.

ii) Set a threshold $N_T$. When $N(R_s)$, the number of literals in $R_s$, is greater than a pre-defined threshold $N_T$ and $D_k^- \neq \emptyset$, then the construction of ETC stops at step *k*.

**Example 4.2** Assume a clause set $S=\{C_1, C_2, C_3, C_4, C_5, C_6, C_7, C_8, C_9, C_{10}\}$ in propositional logic, where $C_1=x_1$, $C_2=x_2$, $C_3=\sim x_1 \vee \sim x_2 \vee x_3$, $C_4=\sim x_1 \vee x_4$, $C_5=\sim x_4 \vee x_5$, $C_6=\sim x_3 \vee \sim x_5$, $C_7=\sim x_3 \vee x_7$, $C_8=\sim x_5 \vee \sim x_7$, $C_9=\sim x_3 \vee \sim x_4$, $C_{10}=\sim x_2 \vee \sim x_7$ with $x_1, x_2, x_3, x_4, x_5, x_6, x_7$ being propositional variables. We can have the following three approaches for constructing ETC.

**Approach 1.**

Table 4.2 ETC constructed based on approach 1

| $C_3$ | $C_7$ | $C_{10}$ | $C_2$ | $C_1$ |
|---|---|---|---|---|
| $x_3$ | $\sim x_3$ | | | |
| | $x_7$ | $\sim x_7$ | | |
| $\sim x_2$ | | $\sim x_2$ | $x_2$ | |
| $\sim x_1$ | | | | $x_1$ |

**Approach 2.**

Table 4.3 ETC constructed based on approach 2

| $C_6$ | $C_5$ | $C_4$ | $C_3$ | $C_2$ | $C_1$ |
|---|---|---|---|---|---|
| $\sim x_5$ | $x_5$ | | | | |
| | $\sim x_4$ | $x_4$ | | | |
| $\sim x_3$ | | | $x_3$ | | |
| | | | $\sim x_2$ | $x_2$ | |
| | | $\sim x_1$ | $\sim x_1$ | | $x_1$ |

**Approach 3.**

Table 4.4 ETC constructed based on approach 3

| $C_3$ | $C_6$ | $C_5$ | $C_4$ | $C_2$ | $C_1$ |
|---|---|---|---|---|---|
| $x_3$ | $\sim x_3$ | | | | |
| | $\sim x_5$ | $x_5$ | | | |
| | | $\sim x_4$ | $x_4$ | | |
| $\sim x_2$ | | | | $x_2$ | |
| $\sim x_1$ | | | $\sim x_1$ | | $x_1$ |

It can be seen from Tables 4.2 – 4.4 that all the above three approaches are able to generate ETCs with no literal above the main boundary line, i.e., the CSC part being empty, and therefore it can be concluded that the original clause set *S* is unsatisfiable.

## 4.4 Extended Triangular Deduction in Propositional Logic

It is usually impossible or difficult in reality to determine the logical property of the clause set based on only one ETC construction as shown in Example 4.2. It is often necessary to add the generated CSC to the initial clause set and start the new round ETC construction and separation several times, which is actually an ETC based automated deduction process as defined as follows.

**Definition 4.2** Given a clause set $S =\{C_1, C_2, \ldots, C_m\}$ in propositional logic, the clause sequence $\Phi_1, \Phi_2, \cdots, \Phi_t$ is called a contradiction separation deduction based on ETM in propositional logic, Extended Triangular deduction in short, from $S$ to $\Phi_t$, if the following conditions hold for $i=1, 2, \ldots, t$.

(1) $\Phi_i \in S$, or

(2) There is $r_1, r_2, \cdots, r_{k_i} < i$, such that $\Phi_i = \mathbb{C}_{k_i}(\Phi_{r_1}, \Phi_{r_1}, \ldots, \Phi_{r_{k_i}})$, where the CSC $\mathbb{C}_{k_i}(\Phi_{r_1}, \Phi_{r_1}, \ldots, \Phi_{r_{k_i}})$ is obtained based on Extended Triangular method.

The general steps of Extended Triangular deduction in propositional logic can be described as follows.

**Step 1. Construct ETC and generate the CSC**

Select the clauses $D_1, D_2, \ldots, D_k$ $(k \geq 2)$ successively from the clause set $S =\{C_1, C_2,\ldots, C_m\}$ $(m \geq 2)$ in propositional logic, based on the method described in subsection 4.3, to construct the ETC and generate the CSC as $R_s = \bigvee_{i=1}^{k} D_i^+$.

**Step 2. Determine the logical property of the clause sets**

Consider the following cases for the generated CSC $R_s = \bigvee_{i=1}^{k} D_i^+$ in Step 1.

**i) When $k \geq m$,**

A. If $R_s = \emptyset$, then $S$ is unsatisfiable.

B. If $R_s \neq \emptyset$, $D_k^+ \neq \emptyset$, then all the clauses in $S$ are included the ETC constructed above, then $S$ is satisfiable. For any $x \in D_k^+$, then $x, x_{k-1}, \ldots, x_i, \ldots, x_1$ is a set of satisfiable instances of $S$ based on Lemma 4.2.

**ii) When $k < m$,**

A. If $R_s = \emptyset$, then $S$ is unsatisfiable.

B. Otherwise, the obtained ETC CSC $R_s = \bigvee_{i=1}^{k} D_i^+$ is added to the original clause set $S$, and let $S' = S \cup \{R_s\}$, then apply the next round Extended Triangular deduction to $S'$ and turn to Step (1) until obtaining a CSC whose property can be determined (the conclusion that the clause set is undecidable might also be obtained for real problems when the predefined stop conditions are reached).

Table 4.4 The deduction result of the first ETC

| $C_3$ | $C_2$ | $C_4$ | $C_7$ | $C_6$ | $C_1$ | $C_8$ |
|---|---|---|---|---|---|---|
| $\sim x_2$ | $x_2$ | | | $x_4$ | | |
| | $x_5$ | $\sim x_5$ | | | | |
| $x_1$ | | $x_1$ | $\sim x_1$ | | | |
| | | | $\sim x_6$ | $x_6$ | | |
| | $x_3$ | | | $x_3$ | $\sim x_3$ | |
| | | | | | $\sim x_7$ | $x_7$ |

**Example 4.3**  Assume a clause set $S=\{C_1, C_2, C_3, C_4, C_5, C_6, C_7, C_8\}$ in propositional logic, where $C_1=\sim x_3 \vee \sim x_7$, $C_2=x_2 \vee x_5 \vee x_3$, $C_3=x_1 \vee \sim x_2$, $C_4=x_1 \vee \sim x_5$, $C_5=\sim x_1 \vee \sim x_4$, $C_6=x_6 \vee x_4 \vee x_3$, $C_7=\sim x_6 \vee \sim x_1$, $C_8=x_7$, with $x_1, x_2, x_3, x_4, x_5, x_6, x_7$ being propositional variables. We can then have the following ETC based deduction as shown in Table 4.4.

The ETC obtained in the first round $C_9=R_1=x_4$ is then added to the original clause set $S$, and the new clause set is denoted as $S' = S \cup \{C_9\}$. Apply the ETM to $S'$, and the result is shown in Table 4.5.

Table 4.5 The deduction result of the second ETC

| $C_1$ | $C_2$ | $C_3$ | $C_4$ | $C_5$ | $C_9$ | $C_8$ |
|---|---|---|---|---|---|---|
| $\sim x_3$ | $x_3$ | | | | | |
| | $x_2$ | $\sim x_2$ | | | | |
| | $x_5$ | | $\sim x_5$ | | | |
| | | | $x_1$ | $x_1$ | $\sim x_1$ | |
| | | | | | $\sim x_4$ | $x_4$ | |
| $\sim x_7$ | | | | | | $x_7$ |

Empty CSC is obtained in the second round ETC construction, and so the process stops with the conclusion being drawn that the original clause set $S$ is unsatisfiable.

**4.5 Soundness and Completeness of the Extended Triangular Deduction Method in Propositional Logic**

As the Extended Triangular deduction is a special case of contradiction separation based automated deduction, the soundness theorem follows directly from Theorem 3.1.

**Theorem 4.2 (Soundness Theorem in Propositional Logic)** Assume that $S = \{C_1, C_2, …, C_m\}$ is a clause set in propositional logic, and $\Phi_1, \Phi_2,…,\Phi_t$ is an Extended Triangular deduction sequence from $S$ to $\Phi_t$. If $\Phi_t$ is an empty clause, then $S$ is unsatisfiable.

**Theorem 4.3 (Completeness Theorem in Propositional Logic)** If a clause set $S = \{C_1,…, C_m\}$ in propositional logic is unsatisfiable, then there is an Extended Triangular deduction from $S$ to empty clause.

**Proof.** Let $n$ be the number of literals in $S = \{C_1, …, C_m\}$, and $K(S) = n - m \geq 0$.

(1) When $K(S) = 0$, it means that all the clauses in $S$, i.e., $C_1, …, C_m$, are unit literals. Therefore, there is at least one complementary pair of literals in $C_1, …, C_m$ given that $S$ is unsatisfiable, and then there exists Extended Triangular deduction from $S$ to empty clause by simply selecting the corresponding two unit clauses to construct the ETC.

(2) Suppose that there is Extended Triangular deduction from $S$ to empty clause when $K(S) \leq t$ ($t>0$), and we need to prove that the conclusion holds for $K(S) = t + 1$ as well.

There is at least one clause that contains more than one literal, which is denoted as $C_1 = p + C$, where $p$ is a literal and $C$ is not empty. Then, $S$ can be divided into two clause sets, $S_1=\{C, C_2, …, C_m\}$ and $S_2=\{p, C_2, …, C_m\}$, and $S$ is unsatisfiable if and only if both $S_1$ and $S_2$ are unsatisfiable.

i) For $S_1$, $K(S_1) = K(S) - 1 = t$, and therefore, there exists Extended Triangular deduction, denoted as $D_1$, from $S_1$ to empty clause based on the induction assumption. Add literal $p$ to all the clauses $C$ appearing in $D_1$, i.e., restore $C$ back to $C_1$, then we obtain a new Extended Triangular deduction $D_{1p}$, where the deduction result will be empty clause or a unit clause containing $p$. If the deduction result of $D_{1p}$ is empty clause, then the induction conclusion holds. If the deduction result is the unit clause containing $p$, then turn to (ii).

ii) For $S_2$, $K(S_2) = K(S) - |C| \leq t$, and therefore, there exists Extended Triangular deduction, denoted as $D_2$, from $S_2$ to empty clause based on the induction assumption. Link $D_{1p}$ obtained from step (i) and $D_2$, and then we can obtain an Extended Triangular deduction from $S$ to empty clause.

**4.6 Some Strategies for Extended Triangular Deduction in Propositional Logic**

For the implementation of Extended Triangular deduction method, we have the following basic strategies.

(1) The general principle: Try to make the leftover literals above the main boundary line as less as possible if it is for unsatisfiability determination, and try to make the leftover literals above the main boundary line as many as possible if it is for satisfiability determination.

(2) There must be no complementary pair of literals in the leftover literals above the main boundary line. If there is, then stop the construction process.

(3) It is in general not possible to change the order of the clauses involving in the same round ETC construction. A new ETC might be formed, and so a new CSC will be generated, if the clause order is changed.

The above three strategies are just the basic requirements for ETM, along with the requirement according to Def. 4.1 that there must be no complementary pair of literals in the main boundary line.

Based on the definition and the above analysis, we can see that the main boundary line literals play an important role on ETC construction, and the process of ETC construction is actually the process of selecting main boundary line literals, because these literals determine the literals inside the ETC, and in turn determine the contradiction. Furthermore, the main boundary line literals have duality.

i) If all of the clauses are used in the ETC with some might being used repeatedly, then a set of satisfiability instances can be obtained based on the Extended Triangular deduction steps discussed in Subsection 3.4.

ii) If a contradiction is already formed during the ETC construction process, then literals in the main boundary line are decisive literals for this contradiction.

Therefore, we have the following strategies on the main boundary line literal selection specifically.

(1) Literals in unit clauses are 'must-select' clauses, and these literals are usually put in the main boundary line firstly. For satisfiability case, these unit clauses must be satisfied, while for unsatisfiability case, these unit clauses play the role on pulling the complementary literals into the contradiction. After that, the subsequent 'must-select' literals that are determined by the current main boundary line literals and the corresponding clauses will be selected iteratively, and this is an iteration process because the main boundary line is updating dynamically.

(2) It is preferred to select the leftover literals as new main boundary line literal during the ETC construction.

Actually, if literal *y* is already left above the main boundary line, and the subsequently selected clause contains *y*, then it is better to select *y* as the new main boundary line literal. By doing this, the possible literal ∼*y* appearing in the subsequent clauses will be pulled into the contradiction, and then will not contribute to the CSC to form a tautology, although it will not be able to remove literal *y* from the CSC.

On the other hand, the aims of constructing ETC are different for satisfiable and unsatisfiable clause sets. Actually, the aim is mainly looking for satisfiable instances for satisfiable clause sets, while determining the unsatisfiability for unsatisfiable clause sets. We can then propose corresponding clause and literal selection methods and strategies for different cases so as to improve the efficiency of ETM.

(1) When aiming at looking for satisfiable instances, we can select the main boundary line literals based on the following priority strategies.

   A. The main boundary line literals can be repeatedly used, but may come from different clauses, because it helps to include more clauses in ETC, and therefore helps to find the satisfiable instances.

   B. Select the literals appearing more times in *S* successively from right to left, or,

   C. Select the literals whose complementary literal appearing less times in *S* successively from right to left. It is then hopefully that the literals inside the contradiction are as less as possible, while leftover literals are as many as possible, which leaves more opportunities for constructing satisfiable instances.

   D. For the literals with the same counting number, we can just put them on the main boundary line without further ordering.

(2) When aiming at determining unsatisfiability, we can select the main boundary line literals based on the following priority strategies.

   A. It is unnecessary to use the main boundary line literals repeatedly, because the existing main boundary line literals can already pull the corresponding literals into the contradiction, and the repeatedly used literals cannot make more contributions on pulling literals.

   B. Select literals whose complementary literal appearing more times in *S* successively from right to left, so that the literals inside the contradiction are as many as possible, while the literals in the CSC are as less as possible, which helps on obtaining empty clause. That is, $D_1$ has only one literal $x_1$, and the number of clauses containing ∼$x_1$ is the greatest in *S*. $D_i - (\{\sim x_{i-1}, \ldots, \sim x_1\} \cup \{x_{i-1}, \ldots, x_1\})$ (*i*>1) has only one literal $x_i$, and the number of clauses containing ∼$x_i$ is the greatest in *S*. Or,

   C. Select the literals appearing more times in *S* successively from right to left, i.e., $D_1$ contains the least number of literals in *S*, and $D_i - (\{\sim x_{i-1}, \ldots, \sim x_1\} \cup \{x_{i-1}, \ldots, x_1\} \cup \bigcup_{j=1}^{i-1} D^+)$ (*i*>1) has the least number of literals correspondingly.

   D. For the literals with the same counting number, we can just put them on the main boundary line without further ordering.

# 5. ETC Based Contradiction Separation Automated Deduction Method in First-Order Logic

**4.1 ETC in First-Order Logic**

Given clause set $S =\{C_1, C_2, \ldots, C_m\}$ ($m\geq 2$) in first-order logic, the form of ETC is illustrated in Fig. 5.1, which takes the similar form as that in propositional logic as shown in Fig. 4.2. There is just one additional requirement that the conditions for Extended Triangular method are applied to the literals and clauses after unification under certain substitution $\sigma$. Actually, we have the following definition.

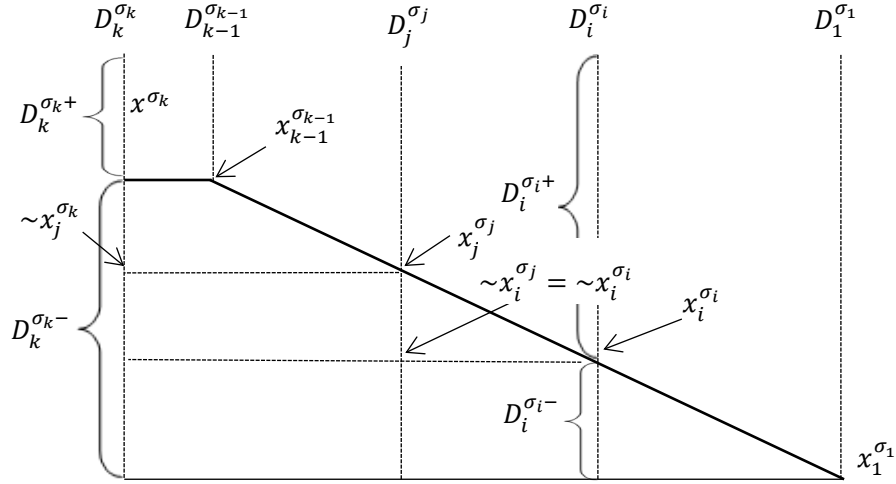

Fig. 5.1 Illustration figure of ETC in first-order logic

Each row in Fig. 5.1, contains the selected literal, after certain substitution, $x_i^{\sigma_i}$ itself and possible complementary literals $\sim x_i^{\sigma_i}$ to the selected literal $x_i^{\sigma_i} \in D_i^{\sigma_i}$, $i=1, 2, \ldots, k-1$ ($k\geq 2$). $x_i^{\sigma_i} \in D_i^{\sigma_i-} \subseteq \{\sim x_1^{\sigma_1}, \cdots, \sim x_{i-1}^{\sigma_{i-1}}, x_i^{\sigma_i}\}$, for $i=1, 2, \ldots, k-1$ ($k\geq 2$), and $\emptyset \neq D_k^{\sigma_k-} \subseteq \{\sim x_1^{\sigma_1}, \cdots, \sim x_{k-1}^{\sigma_{k-1}}\}$.

**Definition 5.1** Assume a clause set $S =\{C_1, C_2, \ldots, C_m\}$ ($m \geq 2$) in first-order logic, where $D_1, D_2, \ldots, D_k$ ($k\geq 2$) are the clauses selected for contradiction construction in one deduction step. The constructed $\bigwedge_{i=1}^{k} D_i^{\sigma_i-}$ is called an ETC in first-order logic, if the following conditions hold.

(1) $D_1, D_2, \ldots, D_k$ share no same variables, and if there are same variables, a rename substitution will be applied to make them different.

(2) For any $D_i$, $i=1, 2, \ldots, k$, a substitution $\sigma_i$ can be applied to $D_i$, ($\sigma_i$ could be an empty substitution) and the same literals merged after substitution, denoted as $D_i^{\sigma_i}$, and $D_i^{\sigma_i}$ 为 can be partitioned into two sub-clauses $D_i^{\sigma_i-}$ and $D_i^{\sigma_i+}$, i.e., $D_i^{\sigma_i} = D_i^{\sigma_i-} \vee D_i^{\sigma_i+}$, such that

i) $D_i^{\sigma_i} = D_i^{\sigma_i-} \vee D_i^{\sigma_i+}$, where $D_i^{\sigma_i-}$ and $D_i^{\sigma_i+}$ do not share the same literal, $D_i^{\sigma_i+}$ can be an empty clause, $D_i^{\sigma_i-}$ cannot be an empty clause, especially, $D_k^{\sigma_k-} \neq \emptyset$; moreover,

ii) There exists literal $x_j \in D_j$ ($j=1, \ldots, k-1$), and there is no complementary pair of literals among $x_1^{\sigma_1}, \cdots, x_{k-1}^{\sigma_{k-1}}$, such that $\emptyset \neq D_k^{\sigma_k-} \subseteq \{\sim x_j^{\sigma_j} | j = 1, \cdots, k-1\}$, $D_i^{\sigma_i-} = \{x_i^{\sigma_i}\} \cup D_i^{\sigma_i 0}$, $D_i^{\sigma_i 0} \subseteq \{\sim x_t^{\sigma_i} | t = 1, \ldots, i-1\}$ ($i=2, \ldots, k-1$), and $D_1^{\sigma_1-} = \{x_1\}$.

Based on Theorem 4.1 in propositional logic, we can see that the ETC defined above is a standard contradiction as stated in the following theorem.

**Theorem 5.1** The ETC defined in Def 4.1, i.e., $\bigwedge_{i=1}^{k} D_i^{\sigma_i -}$, is a standard contradiction.

The disjunction of the literals above the ETC, excluding the main boundary line, i.e., $\bigvee_{i=1}^{k} D_i^{\sigma_i +}$, is the corresponding contradiction separation clause (CSC).

**Example 5.1** Suppose that $S=\{C_1, C_2, C_3, C_4, C_5, C_6, C_7\}$ is a clause set in first-order logic, where $C_1=\sim P_1(x_{11})\vee P_2(x_{11})$, $C_2=\sim P_1(x_{21})\vee P_3(x_{21})$, $C_3=\sim P_3(x_{31})\vee P_4(x_{31})\vee P_5(x_{31})$, $C_4=\sim P_4(x_{41})\vee P_3(f(x_{41}))$, $C_5=P_1(x_{51})$, $C_6=\sim P_5(x_{61})$, $C_7=\sim P_3(f(x_{71}))$, with $P_i$ ($i=1, 2, \ldots, 5$) being predicate symbols, $x_{11}, x_{21}, \ldots, x_{71}$ being variable symbols, $f$ being function symbol. We can then choose substitution $\sigma_i$ ($i=1, \ldots, 7$), to construct the following ETC as shown in Table 5.1, where the substitutions are $\sigma_3 = \emptyset$, $\sigma_4 = x_{31}/x_{41}$, $\sigma_6 = \{x_{31}/x_{61}\}$, and the corresponding CSC is $P_3(f(x_{31}))\vee\sim P_3(x_{31})$.

Table 5.1 1The constructed ETC in Example 5.1

| $C_4^{\sigma_4}$ | $C_3^{\sigma_3}$ | $C_6^{\sigma_6}$ |
|---|---|---|
| $P_3(f(x_{31}))$ | $\sim P_3(x_{31})$ | |
| $\sim P_4(x_{31})$ | $P_4(x_{31})$ | |
| | $P_5(x_{31})$ | $\sim P_5(x_{31})$ |

**4.2 ETC Construction Method in First-Order Logic**

Given a clause set $S =\{C_1, C_2, \ldots, C_m\}$ ($m\geq 2$) in first-order logic, we can select clauses $D_1, D_2, \ldots, D_k$ ($k\geq 2$) from $S$ and the corresponding main boundary line literals successively based on the following steps, to construct the ETC.

**Step 0. Preprocessing**

0.1 Apply deletion strategy to $S$ so as to delete the redundant clauses, and denote the new clause set still as $S$ for simplicity.

0.2 Apply rename substitution to $S$ so that there is no same variables among the clauses in $S$.

**Step 1. Selection of the first clause $D_1$ and the main boundary line literal $x_1^{\sigma_1}$**

Select clause $D_1$ (usually with least number of literals) from the clause set $S$ and the corresponding literal $x_1 \in D_1$, and apply substitution $\sigma_1$, denote $D_1^{\sigma_1 -} = \{x_1^{\sigma_1}\}$, $D_1^{\sigma_1 +} = D_1^{\sigma_1} - D_1^{\sigma_1 -}$, where the literals in $D_1^{\sigma_1 -}$ and $D_1^{\sigma_1 +}$ are in disjunction form, i.e., $D_1^{\sigma_1 -}$ and $D_1^{\sigma_1 +}$ form two sub-clauses. For simplicity, we denote $D_1 = D_1^{\sigma_1}$.

…

**Step $i$. Selection of clause $D_i$ ($2\leq i\leq k-1$) and the main boundary line literal $x_i^{\sigma_i}$**

Select clause $D_i$ from the clause set $S$ and the corresponding literal $x_i \in D_i$,

(1) For $x_j$ ($j \in \{1,2,\cdots,i-1\}$), if there is $y \in D_i$ and substitution $\sigma_i$ such that $x_j^{\sigma_j} = \sim y^{\sigma_i}$, then apply substitution $\sigma_i$ to clause $D_i$, merge the same literals in $D_i^{\sigma_i}$, and put $y^{\sigma_i}$ at the same row with $x_j^{\sigma_j}$.

(2) If there is no such literal $y$ as described in (1), then let $\sigma_i = \emptyset$ (empty substitution).

Denote $D_i^{\sigma_i-} = \{x_i^{\sigma_i}\} \cup D_i^{\sigma_i 0}$, $D_i^{\sigma_i 0} \subseteq \{\sim x_t^{\sigma_t} | t = 1, \ldots, i-1\}$, $i=2, \ldots, k\text{-}1$, $D_i^{\sigma_i+} = D_i^{\sigma_i} - D_i^{\sigma_i-}$, where the literals in $D_i^{\sigma_i-}$ and $D_i^{\sigma_i+}$ are in disjunction. $D_i^{\sigma_i-}$ is the set of literals of $D_i^{\sigma_i}$ inside the ETC, and $D_i^{\sigma_i 0}$ is the leftover literals by subtracting $\{x_i^{\sigma_i}\}$ from $D_i^{\sigma_i-}$, i.e., $D_i^{\sigma_i-}$ and $D_i^{\sigma_i+}$ form two sub-clauses.

…

**Step $k$. Selection of clause $D_k$ (the construction process stops at step $k$)**

Select clause $D_k$ from the clause set $S$ and let $\emptyset \neq D_k^{\sigma_k-} \subseteq \{\sim x_j^{\sigma_j} | j = 1, \cdots, k-1\}$, $D_k^{\sigma_k+} = D_k^{\sigma_k} - D_k^{\sigma_k-}$, where the literals in $D_k^{\sigma_k-}$ and $D_k^{\sigma_k+}$ are in disjunction form

This round of ETC stops at Step $k$, and the obtained CSC is $R_s = \bigvee_{i=1}^{k} D_i^{\sigma_i+}$.

**Remark 5.1** (1) The selection strategies of clause $D_i$, literal $x_i$ ($i$=1, 2,…, $k$-1), and the last clause $D_k$, and the stopping conditions for ETC construction are similar as that in Remark 4.4 for propositional logic, while just need to consider the corresponding substitutions and merging the same literals.

(2) The selection of substitution $\sigma_i$ in step $i$.

i) Select substitution $\sigma_i$ at priority if it can make more literals in $D_i^{\sigma_i}$ have complementary literals in the existing main boundary line.

ii) On the other hand, the selection of substitution $\sigma_i$ can also be made by considering the literal in $D_i$ complementary to $x_j$ for $j \in \{1, 2, \cdots, i-1\}$.

**Example 5.2** ([Wang and Zhou 2017]) Given a clause set $S = \{C_1, C_2, \ldots, C_7\}$ in first-order logic, where $C_1=P_1(a)$, $C_2=\sim P_2(a, b)$, $C_3=P_3(a, f(c), f(b))$, $C_4=P_3(x_1, x_1, f(x_1))$, $C_5=\sim P_3(x_2, x_3, x_4) \vee P_3(x_3, x_2, x_4)$, $C_6=\sim P_3(x_5, x_6, x_7) \vee P_2(x_5, x_7)$, $C_7=\sim P_1(x_8) \vee \sim P_3(x_9, x_{10}, x_{11}) \vee \sim P_2(x_8, x_{11}) \vee P_2(x_8, x_9) \vee P_2(x_8, x_{10})$ with $P_i$ ($i$=1, 2, 3) being predicate symbols, $x_i$ ($i$=1,…,11) being variable symbols, $a$, $b$, $c$ being constant symbols, $f$ being function symbol. We can then select substitution $\sigma_i$ ($i$=1, …, 7), and construct the ETC as shown in Table 5.2, where the substitutions are $\sigma_1 = \emptyset$, $\sigma_2 = \emptyset$, $\sigma_3 = \emptyset$, $\sigma_4 = \{b/x_1\}$, $\sigma_6 = \{a/x_5, f(c)/x_6, f(b)/x_7\}$, $\sigma_7 = \{a/x_8, b/x_9, b/x_{10}, f(b)/x_{11}\}$.

Table 5.2 The ETC constructed for Example 5.2

| $C_7^{\sigma_7}$ | $C_6^{\sigma_6}$ | $C_4^{\sigma_4}$ | $C_3^{\sigma_3}$ | $C_2^{\sigma_2}$ | $C_1^{\sigma_1}$ |
|---|---|---|---|---|---|
| $\sim P_2(a, f(b))$ | $P_2(a, f(b))$ | | | | |
| $\sim P_3(b, b, f(b))$ | | $P_3(b, b, f(b))$ | | | |
| | $\sim P_3(a, f(c), f(b))$ | | $P_3(a, f(c), f(b))$ | | |
| $P_2(a, b)$ | | | | $\sim P_2(a, b)$ | |
| $P_2(a, b)$ | | | | | |
| $\sim P_1(a)$ | | | | | $P_1(a)$ |

**Remark 5.2** Example 5.2 shows that the unsatisfiability of the clause set is determined by constructing only one ETC, while it is impossible to obtain this result by constructing only one Standard Triangle. It can also be concluded that clause $C_5$ is a redundant clause in $S$ because it is not involved in the only ETC.

### 5.3 Soundness and Completeness of the Extended Triangular Deduction Method in First-Order Logic

Similar as that in first-order logic, it is usually impossible or difficult in reality to determine the logical property of the clause set based on only one ETC construction as shown in Example 5.2, and it is usually necessary to put the generated CSC into the initial clause set and start next round ETC construction and separation several times, which is actually an ETC based automated deduction process as defined in Def. 5.2.

**Definition 5.2** Given a clause set $S = \{C_1, C_2, \ldots, C_m\}$ ($m \geq 2$) in first-order logic, the clause sequence $\Phi_1, \Phi_2, \cdots, \Phi_t$ is called a contradiction separation deduction based on Extended Triangular method in first-order logic, Extended Triangular deduction in short, from $S$ to $\Phi_t$, if the following conditions hold for $i=1, 2, \ldots, t$.

(1) $\Phi_i \in S$, or

(2) There is $r_1, r_2, \cdots, r_{k_i} < i$, and substitutions $\sigma_{r_1}, \sigma_{r_2}, \cdots, \sigma_{r_{k_i}}$, such that $\Phi_i = \mathbb{C}_m^s\left(\Phi_{r_1}^{\sigma_{r_1}}, \Phi_{r_2}^{\sigma_{r_2}}, \cdots, \Phi_{r_{k_i}}^{\sigma_{r_{k_i}}}\right)$, where the CSC $\mathbb{C}_{k_i}^s\left(\Phi_{r_1}^{\sigma_{r_1}}, \Phi_{r_2}^{\sigma_{r_2}}, \cdots, \Phi_{r_{k_i}}^{\sigma_{r_{k_i}}}\right)$ is generated based on Extended Triangular method.

The general steps of Extended Triangular deduction in first-order logic can be described as follows.

**Step 1. Construct ETC and generate the CSC**

Select the clauses $D_1, D_2, \ldots, D_k$ ($k \geq 2$) successively from the clause set $S = \{C_1, C_2, \ldots, C_m\}$ ($m \geq 2$) in first-order logic, based on the method described in subsection 5.2, to construct the ETC and generate the CSC as $R_s = \bigvee_{i=1}^{k} D_i^{\sigma_i+}$.

**Step 2. Determine the logical property of the clause sets based on the CSC**

Consider the following cases for the CSC $R_s = \bigvee_{i=1}^{k} D_i^{\sigma_i+}$ generated in Step 1.

(1) If $R_s = \emptyset$, then $S$ is unsatisfiable.

(2) Otherwise, the obtained ETC is added to the original clause set $S$, and let $S' = S \cup \{R_s\}$, then apply the next round Extended Triangular deduction to $S'$ and turn to Step (1) until obtaining a CSC whose property can be determined (the conclusion that the clause set is undecidable might also be obtained for real problems when the predefined stop conditions are reached).

We can then provide the soundness and completeness theorems for Extended Triangular deduction method. As the Extended Triangular deduction is a special case of contradiction separation based automated deduction, the soundness theorem follows directly from Theorem 3.3. That is, Extended Triangular deduction method in first-order logic is sound because the contradiction separation based automated deduction is sound in first-order logic.

**Theorem 5.2 (Soundness Theorem in First-Order Logic)** Assume that $S = \{C_1, C_2, \ldots, C_m\}$ is a clause

set in first-order logic, and $\Phi_1, \Phi_2, \ldots, \Phi_t$ is an Extended Triangular deduction sequence from $S$ to $\Phi_t$. If $\Phi_t$ is an empty clause, then $S$ is unsatisfiable.

According to the proof of the Completeness Theorem of S-CS based deduction in first-order logic, i.e., Theorem 3.4, there is no special requirements on the selected contradiction [Xu *et al*. 2018]. It means that all the contradictions in the proof can be replaced by ETCs, and so the Completeness Theorem for Extended Triangular deduction in first-order logic hold still.

**Theorem 5.3** (**Completeness Theorem in First-Order Logic**) If a clause set $S = \{C_1, C_2, \ldots, C_m\}$ in first-order logic is unsatisfiable, then there is an Extended Triangular deduction from $S$ to empty clause.

**Remark 5.3** In fact, as binary resolution deduction is a special case of Extended Triangular deduction where each contradiction has exactly two clauses, and the binary resolution deduction is complete, so the Extended Triangular deduction, both in propositional logic and first-order logic, is also complete. Furthermore, it can be seen from Example 4.3 and the following Example 5.3 that there is Extended Triangular deduction with contradiction containing more than two clauses, which shows that there does exist Extended Triangular deduction which is not binary resolution deduction.

**Example 5.3** (Liu 1994)

Given a clause set $S = \{C_1, C_2, C_3, C_4, C_5, C_6, C_7\}$ in first-order logic,

$C_1 = \sim P_1(x_{11}, x_{12}, x_{13}) \vee \sim P_2(x_{11}, x_{13})$, $C_2 = P_1(x_{22}, x_{21}, x_{23}) \vee \sim P_1(x_{21}, x_{22}, x_{23})$,

$C_3 = P_2(x_{31}, x_{34}) \vee \sim P_3(x_{31}) \vee \sim P_1(x_{32}, x_{33}, x_{34}) \vee \sim P_2(x_{31}, x_{32}) \vee \sim P_2(x_{31}, x_{33})$,

$C_4 = P_1(x_{41}, x_{41}, f_1(x_{41}))$, $C_5 = P_1(a_1, f_1(a_1), f_1(a_3))$, $C_6 = P_3(a_1)$, $C_7 = P_2(a_1, a_3)$,

where $P_i$ ($i = 1, 2, 3$) are predicate symbols, $x_j$ ($j = 11, \ldots, 41$) are variable symbols, $a_k$ ($k = 1, 2, 3$) are constant symbols, and $f_1$ is a function symbol. We can then have the following two Extended Triangular deductions. Table 5.3 shows the first constructed ETC, where the substitutions are $\sigma_{11} = \{a_1/x_{11}, a_3/x_{13}\}$, $\sigma_{21} = \{a_1/x_{21}, f_1(a_1)/x_{22}, f_1(a_3)/x_{23}\}$, $\sigma_{31} = \{a_1/x_{31}, a_3/x_{32}, a_3/x_{33}, f_1(a_3)/x_{34}\}$, $\sigma_{41} = \{a_3/x_{41}\}$, $\sigma_{51} = \emptyset$, $\sigma_{61} = \emptyset$, $\sigma_{71} = \emptyset$.

Table 5.3 The deduction result of the first round Extended Triangular deduction

| $C_1^{\sigma_{11}}$ | $C_2^{\sigma_{21}}$ | $C_3^{\sigma_{31}}$ | $C_4^{\sigma_{41}}$ | $C_5^{\sigma_{51}}$ | $C_7^{\sigma_{71}}$ | $C_6^{\sigma_{61}}$ |
|---|---|---|---|---|---|---|
| | | $P_2(a_1, f_1(a_3))$ | | | | |
| $\sim P_1(a_1, x_{12}, a_3)$ | | | | | | |
| | $P_1(f_1(a_1), a_1, f_1(a_3))$ | | | | | |
| | | $\sim P_1(a_3, a_3, f_1(a_3))$ | $P_1(a_3, a_3, f_1(a_3))$ | | | |
| | $\sim P_1(a_1, f_1(a_1), f_1(a_3))$ | | | $P_1(a_1, f_1(a_1), f_1(a_3))$ | | |
| $\sim P_2(a_1, a_3)$ | | $\sim P_2(a_1, a_3)$ | | | $P_2(a_1, a_3)$ | |
| | | $\sim P_2(a_1, a_3)$ | | | | |
| | | $\sim P_3(a_1)$ | | | | $P_3(a_1)$ |

Denote the CSC generated by Extended Triangular deduction as shown in Table 5.3 as $C_8 = P_2(a_1, f_1(a_3))$,

and put it back to the original clause set to start a new Extended Triangular deduction, then empty clause is obtained as shown in Table 5.4, where the substitutions are $\sigma_{12} = \{a_1/x_{11}, f_1(a_1)/x_{12}, f_1(a_3)/x_{13}\}$, $\sigma_{52} = \emptyset$, $\sigma_{82} = \emptyset$.

Table 5.4 The deduction result of the second round Extended Triangular deduction

| $C_5^{\sigma_{52}}$ | $C_1^{\sigma_{12}}$ | $C_8^{\sigma_{82}}$ |
|---|---|---|
| $P_1(a_1, f_1(a_1), f_1(a_3))$ | $\sim P_1(a_1, f_1(a_1), f_1(a_3))$ | |
| | $\sim P_2(a_1, f_1(a_3))$ | $P_2(a_1, f_1(a_3))$ |

Actually, we can also obtain the empty clause based on just one round Extended Triangular deduction, as shown in Table 5.5, where $\sigma_{13} = \{a_1/x_{11}, f_1(a_1)/x_{12}, f_1(a_3)/x_{13}\}$, $\sigma_{33} = \{a_1/x_{31}, a_3/x_{32}, a_3/x_{33}, f_1(a_3)/x_{34}\}$, $\sigma_{43} = \{a_3/x_{41}\}$, $\sigma_{53} = \emptyset$, $\sigma_{63} = \emptyset$, $\sigma_{73} = \emptyset$.

Table 5.5 The deduction result based on just one round Extended Triangular deduction

| $C_3^{\sigma_{33}}$ | $C_1^{\sigma_{13}}$ | $C_4^{\sigma_{43}}$ | $C_5^{\sigma_{53}}$ | $C_7^{\sigma_{73}}$ | $C_6^{\sigma_{63}}$ |
|---|---|---|---|---|---|
| $P_2(a_1, f_1(a_3))$ | $\sim P_2(a_1, f_1(a_3))$ | | | | |
| $\sim P_1(a_3, a_3, f_1(a_3))$ | | $P_1(a_3, a_3, f_1(a_3))$ | | | |
| | $\sim P_1(a_1, f_1(a_1), f_1(a_3))$ | | $P_1(a_1, f_1(a_1), f_1(a_3))$ | | |
| $\sim P_2(a_1, a_3)$ | | | | $P_2(a_1, a_3)$ | |
| $\sim P_2(a_1, a_3)$ | | | | | |
| $\sim P_3(a_1)$ | | | | | $P_3(a_1)$ |

**Remark 5.5** Example 5.3 further illustrates that the crucial part of Extended Triangular deduction method is the selection of the clauses contributing to the contradiction and the corresponding main boundary line literals, which decides the deduction efficiency of Extended Triangular method.

**5.4 Some Strategies for Extended Triangular Deduction in First-Order Logic**

The strategies for Extended Triangular deduction method in propositional logic suits Extended Triangular deduction method in first-order logic, under the assumption of substitutions and merging the same literals after substitution are applied, and so we list just the general strategies for substitution as follow.

(1) Substitutions that lead to the generation of redundant clause should not be selected. For example, for clause $C$, if substitution $\sigma$ makes $C^\sigma$ a redundant clause, then $\sigma$ cannot be selected.

(2) (Inverse substitution) During the ETC construction process, substitution $\sigma_i$ is applied to clause $D_i$ so as to form more complementary pairs, and pull more literals into the contradiction. On the other hand, this substitution will sometimes affect the clauses selected prior to $D_i$, and this kind of substitution is called *inverse substitution* that was discussed in (Kovács and Voronkov, 2013).

(3) During the construction of ETC, substitution $\theta$ satisfying the following conditions can be added to certain clauses after the normal clause and literal selection and merging the same literals after substitution as described in subsection 4.2.

i) Substitution $\theta$ is able to make the literals outside the ETC 'fall' into the ETC after substitution.

ii) If there are more than one substitutions that satisfies condition i), then all these substitutions can be applied successively.

iii) Such substitution should be applied to all the literals in the clause.

iv) If substitution $\theta$ affects the existing complementary pair, then inverse substitution should be applied to the corresponding clauses. Try to apply as less as possible this kind of inverse substitution if there are more than two clauses are affected. Actually, for this case, we can choose to select other clauses or literals.

## 6. Relation between Extended Triangular Deduction and Linear Deduction

In Ref. (Kovács and Voronkov, 2013), it has already been stated that the Standard Triangular deduction can be realized by linear deduction. On the other hand, the Extended Triangular deduction usually cannot be realized by linear deduction, as shown in the following example.

**Example 6.1** Given a clause set $S$: $D_1 = \sim x_2 \vee \sim x_5$, $D_2 = \sim x_3 \vee x_2$, $D_3 = x_3 \vee \sim x_5$, $D_4 = x_4 \vee \sim x_3$, $D_5 = x_3 \vee \sim x_1$, $D_6 = x_5 \vee \sim x_4$ in first-order logic.

Apply the Extended Triangular deduction to $S$ as shown in Table 6.1, and we can obtain the CSC as $R = \sim x_1 \vee \sim x_4$.

Table 6.1 The ETC construction process for Example 6.1

| $D_1$ | $D_2$ | $D_3$ | $D_4$ | $D_5$ | $D_6$ |
|---|---|---|---|---|---|
| $\sim x_2$ | $x_2$ | | | | |
| | $\sim x_3$ | $x_3$ | | | |
| | | | $x_4$ | $\sim x_1$ | |
| | | | $\sim x_3$ | $x_3$ | $\sim x_4$ |
| $\sim x_5$ | | $\sim x_5$ | | | $x_5$ |

It can be seen that linear deduction cannot be formed from right to left for $D_1 = \sim x_2 \vee \sim x_5$, $D_2 = \sim x_3 \vee x_2$, $D_3 = x_3 \vee \sim x_5$, $D_4 = x_4 \vee \sim x_3$, $D_5 = x_3 \vee \sim x_1$, $D_6 = x_5 \vee \sim x_4$, as shown in Fig. 5.1.

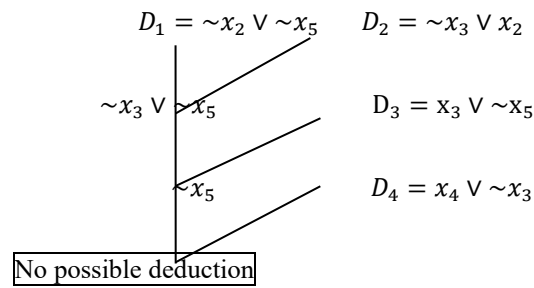

Fig. 6.1 Illustration figure that ETC cannot be realized with linear deduction

On the other hand, we have the following theorem.

**Theorem 6.1** Linear deduction can be realized by Extended Triangular deduction.

**Proof.** Suppose that the linear deduction is as shown in Fig. 6.2.

In Fig. 6.2, $R_{k,k-1} = R(R_k = D_k, D_{k-1}) = (D_k - \{\sim x_{k-1}\}) \vee (D_{k-1} - \{x_{k-1}\})$,

$R_{k,k-1,k-2} = R(R_{k,k-1}, D_{k-2}) = (R_{k,k-1} - \{\sim x_{k-2}\}) \vee (D_{k-2} - \{x_{k-2}\})$,

……

$R_{k,k-1,k-2,\cdots,i} = R(R_{k,k-1,k-2,\cdots,i+1}, D_i) = (R_{k,k-1,k-2,\cdots,i+1} - \{\sim x_i\}) \vee (D_i - \{x_i\})$,

……

$R_{k,k-1,k-2,\cdots,2} = R(R_{k,k-1,k-2,\cdots,3}, D_2) = (R_{k,k-1,k-2,\cdots,3} - \{\sim x_2\}) \vee (D_2 - \{x_2\})$,

$R_{k,k-1,k-2,\cdots,2,1} = R(R_{k,k-1,k-2,\cdots,3,2}, D_1) = (R_{k,k-1,k-2,\cdots,3,2} - \{\sim x_1\}) \vee (D_1 - \{x_1\})$.

(1) If there is no complementary pair of literals among $\{x_{k-1}, x_{k-2}, …, x_i, …, x_2, x_1\}$, then this linear deduction can be realized by the following Extended Triangular deduction.

The bottom part closed by the bold lines in Fig. 6.3 is actually the ETC corresponding to the above linear deduction. When each $\sim x_i$ appears in $D_{i+1}$ ($i=1, …, k-2$), the above contradiction is also a Standard Contradiction.

Fig. 6.2 The linear deduction process

(2) If there is complementary pair among $\{x_{k-1}, x_{k-2}, …, x_i, …, x_2, x_1\}$, suppose that $x_{i0}$ is the first literal having complementary literal in $\{x_{k-1}, x_{k-2}, …, x_i, …, x_2, x_1\}$ from left to right, i.e., there is no complementary pair in $\{x_{k-1}, x_{k-2}, …, x_{i0+1}\}$, then $R_{k,k-1,k-2} = R(R_{k,k-1}, D_{k-2}) = (R_{k,k-1} - \{\sim x_{k-2}\}) \vee (D_{k-1} - \{x_{k-2}\})$,

……,

and

$R_{k,k-1,\cdots,i0+1} = R(R_{k,k-1,\cdots,i0+2}, D_{i0+1}) = (R_{k,k-1,\cdots,i0+2} - \{\sim x_{i0+1}\}) \vee (D_{i0+1} - \{x_{i0+1}\})$

can be realized by ETC.

| | | | | | | | | | | $R_{k,k-1,k-2,...,1}$ | |
|---|---|---|---|---|---|---|---|---|---|---|---|
| | | | | | | | | | $R_{k,k-1,k-2,...,2}-\{\sim x_1\}$ | | |
| | | | | | | | | | . | | |
| | | | | | | | | | . | | |
| | | | | | | | $R_{k,k-1,k-2,...,i-1}-\{\sim x_{i-2}\}$ | | | | |
| | | | | | | $R_{k,k-1,k-2,...,i}-\{\sim x_{i-1}\}$ | | | | | |
| | | | | | $R_{k,k-1,k-2,...,i+1}-\{\sim x_i\}$ | | | | | | |
| | | | . | | | | | | | | |
| | | | . | | | | | | | | |
| | | $R_{k,k-1,k-2}-\{\sim x_{k-3}\}$ | | | | | | | | | |
| $R_{k,k-1}-\{\sim x_{k-2}\}$ | | | | | | | | | | | |
| $D_k$ | $D_{k-1}$ | $D_{k-2}$ | ... | $D_{i+1}$ | $D_i$ | $D_{i-1}$ | ... | $D_2$ | $D_1$ | | |
| $\sim x_{k-1}$ | $x_{k-1}$ | | | | | | | | | | |
| $\exists \sim x_{k-2}$ | | $x_{k-2}$ | | | | | | | | | |
| | | | . | | | | | | | | |
| | | | . | | | | | | | | |
| $\exists \sim x_i$ | | | | | $x_i$ | | | | | | |
| $\exists \sim x_{i-1}$ | | | | | | $x_{i-1}$ | | | | | |
| | | | | | | | . | | | | |
| $\exists \sim x_2$ | | | | | | | | $x_2$ | | | |
| $\exists \sim x_1$ | | | | | | | | | $x_1$ | | |

Fig. 6.3 The ETC corresponding to linear deduction

(3) If there is no complementary pair in $\{x_{i0}, x_{i0-1}, ..., x_i, ..., x_2, x_1\}$, then $R_{k,k-1,k-2,\cdots,i0+1}$,

$R_{k,k-1,k-2,\cdots,i0} = R(R_{k,k-1,k-2,\cdots,i0+1}, D_{i0}) = (R_{k,k-1,k-2,\cdots,i0+1} - \{\sim x_{i0}\}) \vee (D_{i0} - \{x_{i0}\})$,

……,

$R_{k,k-1,k-2,\cdots,i} = R(R_{k,k-1,k-2,\cdots,i+1}, D_i) = (R_{k,k-1,k-2,\cdots,i+1} - \{\sim x_i\}) \vee (D_i - \{x_i\})$,

……,

$R_{k,k-1,k-2,\cdots,2} = R(R_{k,k-1,k-2,\cdots,3}, D_2) = (R_{k,k-1,k-2,\cdots,3} - \{\sim x_2\}) \vee (D_2 - \{x_2\})$,

and $R_{k,k-1,k-2,\cdots,2,1} = R(R_{k,k-1,k-2,\cdots,3,2}, D_1) = (R_{k,k-1,k-2,\cdots,3,2} - \{\sim x_1\}) \vee (D_1 - \{x_1\})$,

can be realized by ETC.

(4) If there is complementary pair among $\{x_{i0}, x_{i0-1}, \ldots, x_i, \ldots, x_2, x_1\}$, then repeat step (2) until the linear deduction can be realized by ETC piecewise.

**Remark 6.1** If the most general unifier $\sigma_{k,k-1,k-2,\cdots,i+1}$ is used in the linear deduction for $R_{k,k-1,k-2,\cdots,i} = R(R_{k,k-1,k-2,\cdots,i+1}, D_i) = (R_{k,k-1,k-2,\cdots,i+1}) - \{\sim x_i\}) \vee (D_i - \{x_i\})$, then we need to replace it with the following equation with $x_i$ being replaced by $x_i^{\sigma_{k,k-1,k-2,\cdots,i+1}}$ under substitution.

$$R_{k,k-1,k-2,\cdots,i} = R\big(R_{k,k-1,k-2,\cdots,i+1}^{\sigma_{k,k-1,k-2,\cdots,i+1}}, D_i^{\sigma_{k,k-1,k-2,\cdots,i+1}}\big)$$

$$= (R_{k,k-1,k-2,\cdots,i+1}^{\sigma_{k,k-1,k-2,\cdots,i+1}} - \{\sim x_i^{\sigma_{k,k-1,k-2,\cdots,i+1}}\}) \vee (D_i^{\sigma_{k,k-1,k-2,\cdots,i+1}} - \{x_i^{\sigma_{k,k-1,k-2,\cdots,i+1}}\}).$$

This shows that linear deduction can be realized by ETC piecewise, while ETC cannot be realized by linear deduction usually. Therefore, linear deduction is a special case of Extended Triangular deduction.

## 7. Experimental Evidence and System Validation

Because the proposed *Extended Triangular Method* (ETM) serves as the **algorithmic realization** of the theoretical CSE framework rather than as a standalone prover, its empirical performance cannot be evaluated in isolation. Actually, the ETM provides the **generalized construction foundation** for a series of CSE–based automated theorem provers developed in recent years, including **CSE, CSE_E, CSI_E, and CSI_Enig**. Each of these systems has been implemented upon, or directly benefited from, the triangular and extended-triangular contradiction construction principles formalized in this paper. Therefore, instead of conducting independent experiments, this section reviews the performance of these systems in international benchmark competitions (e.g., CASC - CADE ATP System Competition 2018-2025), which is legitimate to serve as strong empirical validation of the effectiveness and scalability of the proposed inference mechanism. These systems, developed between 2018 and 2025, collectively demonstrate how the ETM-based contradiction construction translates theoretical innovation into practical performance.

### 7.1 Algorithmic Influence, Naming Lineage and System Evolution

It is worth noting that a preliminary algorithmic realization of contradiction separation, termed the *Standard Extension (SE)* method (Xu *et al.* 2025b), *was previously proposed as the first algorithmic implementation of the 2018 CSE framework*. That study focused on concretely defining how contradictions can be incrementally constructed and extended through complementary literals. In contrast, the present paper generalizes that mechanism into the **Extended Triangular Contradiction Separation Method**, termed ETM, which unifies all known contradiction-construction strategies within a single dynamic framework, and *provides the theoretical and algorithmic foundation*. Accordingly, while the same family of systems (CSE, CSE_E, CSI_E, and CSI_Enig) is referenced for empirical validation, their role here differs: in the earlier work, their competition performance substantiated the *practical effectiveness* of the SE algorithm; in the current study, those results further corroborate the *generality and theoretical soundness* of the Extended Triangular framework that subsumes all prior algorithmic realizations.

The naming of the **CSE family** of systems reflects the historical evolution of the contradiction separation framework from theory to implementation. The prefix **CS** originates from *Contradiction*

*Separation*, the core inference principle that redefines logical contradiction as a cooperative relationship among multiple clauses rather than a pairwise complementarity. The suffix **E** in **CSE** denotes *Extension*, indicating that the **CSE framework** proposed in 2018 extends the basic CS concept into a comprehensive theory of *dynamic multi-clause synergized automated deduction*, complete with formal proofs of soundness and completeness. Building on this foundation, subsequent systems preserved the established "CSE" prefix to emphasize their theoretical continuity. All the above provers along with system description are available via the Systems and Entrants Lists published on the CASC website (2018-2025), but without an explicit published contradiction-construction algorithmic description. Against this backdrop, The present work fills this gap to complete the theoretical–algorithmic bridge.

The following summarize the systems incorporate or extend the ETM and their evolutions:

- **CSE Prover (Baseline) (2018):** Implemented the first dynamic multi-clause reasoning prototype, based on the conceptual principles of contradiction separation but without an explicit published algorithmic description. The system demonstrated the feasibility of contradiction separation as a scalable alternative to binary resolution in first-order logic reasoning.
- **CSE_E (2018–2025)** extends the original CSE prover by integrating the *E* superposition calculus for equality reasoning, creating a hybrid system that combines dynamic multi-clause deduction with robust equational inference. Through this integration, CSE_E leverages the strengths of both paradigms—CSE's dynamic contradiction separation and *E*'s efficient equality handling, achieving higher problem-solving coverage and performance on TPTP and CASC benchmarks. It establishes a bridge between contradiction separation reasoning and established saturation-based proof engines.
- **CSI_E (2024–2025)** is a multi-layer inverse and parallel automated theorem prover built upon the *CSE* framework and integrated with the *E* superposition prover. It extends the CSE paradigm by enabling inverse and parallel contradiction separation across multiple clause layers, dynamically generating sub-goals and unit clauses that enhance *E*'s proof search and equality reasoning. The combination yields improved deductive depth, proof guidance, and efficiency across complex first-order logic problems.
- **CSI_Enigma (2025)** is a multi-layer inverse and parallel first-order automated theorem prover that integrates the *CSE* framework with the *ENIGMA* learning-based guidance mechanism. Building on CSI's dynamic multi-clause reasoning and efficient unit-clause construction, the system employs ENIGMA's neural clause-selection model to prioritize promising inferences and accelerate proof discovery. This synthesis of symbolic and statistical reasoning achieves enhanced scalability, efficiency, and competitiveness in large-scale theorem-proving tasks.

The following small schematic or timeline visualizes the theoretical lineage and practical evolution,

CSE (2018-2025) → CSE_E (2018-2025) → CSI_E (2024 and 2025) → CSI_Enig (2025)

This lineage illustrates a coherent progression from foundational theory (CSE) to and then hybrid and learning-augmented implementations (CSE, CSE_E, CSI_E, CSI_Enig), all grounded in the contradiction separation paradigm and the triangular construction algorithm formalized in this paper.

Although none of these system papers or competition reports explicitly detailed the internal contradiction-construction algorithm, all of them depend on the ETM procedure formalized here. Their verified soundness and superior competition performance (see Section 7.2 below) thus provide indirect but powerful empirical validation of the algorithm's effectiveness.

### 7.2. ATP System Competition-Based Empirical Validation

To demonstrate the empirical impact of the CSE framework and ETM, Table 7.1 summarizes representative results from international competitions **CASC (CADE ATP System Competition)** over the past several years in which systems derived from the proposed methodology have participated.

**Table 7.1** Performance of CSE-family systems in international competitions

| System | Competition (Years) | Category / Domain | Rank / Result (Approx.) | Notable Strengths |
|---|---|---|---|---|
| **CSE (Prover)** | CASC 2018-2025 | FOF | Demonstrated novel multi-clause proofs | Conceptual CSE theory, early ETM prototype |
| **CSE_E** | CASC 2018 – 2025 | FOF & EPR | Consistently Top-5; solved 10–15 % more equality problems than E | Integration of multi-clause CS with superposition |
| **CSI_E** | CASC 2024 and 2025 | FOF and ICU | Cross-prover challenge debut and achieving solved-problem counts comparable to other strong systems | Stronger clause synergy and enhanced inference flexibility, especially in combined or challenge tracks |
| **CSI_Enig** | CASC2025 | FOF and ICU | Comparable to top-performing systems, indicating strong competitiveness | Incorporates ENIGMA-style learning guidance |

Across these evaluations, CSE-based provers demonstrated continuous performance improvement. The CSE_E prover, which integrates the contradiction separation mechanism with the superposition-based *E theorem prover*, has participated in multiple editions of the CASC since 2018. Competition reports consistently show that CSE_E demonstrates higher problem-solving coverage compared with standalone E on equality-heavy and structurally complex benchmarks. In the 2020–2023 CASC FOF and EPR divisions, CSE_E solved a subset of problems that E alone failed to prove, confirming the contribution of CSE-based multi-clause reasoning to overall performance.

### 7.3 Interpretation and Impact

The continuous improvement trend across these systems empirically demonstrates that the Extended Triangular contradiction construction, which generalizes and unifies the core mechanism used in all these provers, effectively enhances both the completeness and efficiency of automated deduction. The empirical evidence from years of international benchmarking indicates that the *ETM* is not merely a theoretical construct but a practically validated algorithmic paradigm. Its capability to unify multi-clause reasoning, equality handling, and learning-based inference guidance under a single constructive mechanism distinguishes it from traditional resolution and hyper-resolution systems. In particular:

- The multi-clause synergized inference enabled larger reasoning steps and reduced redundant derivations.
- The flexible contradiction construction improved goal-directed proof search.
- The integration with external equality and learning modules further expanded scalability.

The consistent advancement in problem-solving rates and proof search efficiency clearly reflects the practical value of the contradiction separation and triangular construction principles developed herein. The empirical results also suggest that ETM provides a strong theoretical and practical foundation for next-generation contradiction separation based automated deduction systems, particularly when integrated with hybrid or learning-augmented architectures such as CSE_E and CSI_Enig.

By formally publishing this algorithm, the present paper provides the **missing algorithmic disclosure** connecting the 2018 theoretical framework to the subsequent implementations that have already proven its effectiveness. The success of those systems in independent evaluations serves as *retrospective experimental validation* for the method now fully defined herein.

Consequently, the ETM can be viewed as the **unified inference-theoretic core** behind these state-of-the-art systems, whose benchmark performance constitutes a credible validation of the proposed framework.

## 8. Conclusions and Future Works

This paper presented the Extended Triangular Contradiction Separation Based Dynamic Automated Deduction Method (ETM in short), a generalized sound and complete framework for multi-clause reasoning in automated deduction. By relaxing the restriction of complementary literal selection, the method achieves greater flexibility and allows real-time adaptation of clause and literal choice during proof construction. Theoretical results confirm its soundness and completeness, while the sustained success of its descendant systems—CSE, CSE_E, CSI_E, and CSI_Enig—in international competitions provides cumulative empirical validation of its effectiveness.

The ETM unifies classical binary-resolution inference and the Standard Triangular method as special cases, establishing a comprehensive inference schema for contradiction-based reasoning. Beyond its formal properties, its compatibility with superposition-based reasoning (in CSE_E) and learning-guided inference (in CSI_Enig) demonstrates that the framework is both theoretically robust and practically extensible, therefor ETM provides an adaptable foundation for integration with learning-guided heuristics and hybrid AI reasoning systems, this also shows the potential for *AI–ATP integration.*

In conclusion, the ETM provides the inference-theoretic core underpinning a new generation of contradiction-separation-based provers. Its theoretical generality, empirical validation, and compatibility with AI-guided reasoning suggest that it can play a central role in advancing dynamic, synergized, and intelligent theorem proving. ETM represents a synthesis of classical logical rigor and modern dynamic reasoning, advancing the frontier of automated deduction toward greater structural awareness, efficiency, and adaptability. Moreover, ETM naturally supports explainable reasoning. The triangular relationships between clauses can be visualized as dependency graphs, enabling post hoc analysis of proof strategies — an increasingly important requirement in interpretable AI (Miller, 2019).

Future research will explore several promising directions:

1) **Learning-Integrated Deduction:** Develop neural and reinforcement learning models to predict clause combinations and optimize contradiction construction dynamically. ETM can be combined with learning-based clause selection (Jakubuv & Urban, 2020) or attention-driven symbolic inference (Polu & Sutskever, 2020), allowing dynamic optimization of the reasoning process.

2) **Cross-Domain Applications:** Extend ETM to non-classical Logic (modal, temporal, higher-order logics, fuzzy logics, and probabilistic logic, enabling reasoning in richer semantic domains, as well as formal verification tasks in software and system reasoning (symbolic model checking or SAT-based verification).

3) **Algorithmic Optimization and Implementation Enhancement:** Design efficient data structures and indexing techniques and substitution mechanisms for large-scale ETM implementations.

4) **Complexity and Redundancy Analysis:** Establish theoretical bounds and heuristic strategies for inference growth, quantifying and minimizing redundancy under multi-clause synergy.

5) **Combination with Semantic Guidance** (semantic tableaux or model elimination): enables more targeted clause expansion and semantic pruning and bridges symbolic and semantic reasoning, offering a unified pathway toward more intelligent and efficient automated deduction.

6) **Hybrid Reasoning Frameworks:** Embed ETM within proof assistants and verification platforms to bridge automated and human-guided deduction.

## Acknowledgements

This work is partially supported by the National Natural Science Foundation of China (Grant No. 61673320, 61976130).